\documentclass[]{article}

\usepackage{amsmath,amsfonts,amssymb,amsthm,amstext,lineno,graphicx,subfigure,color,bbm}
\usepackage{caption}
\usepackage{subfigure}
\usepackage[]{hyperref}
\usepackage{multirow}
\usepackage[latin1]{inputenc}
\usepackage[nolist]{acronym}

\hypersetup{pdftitle={Robust Periocular Recognition By Fusing Sparse Representations of Color and Geometry Information}, colorlinks=true,citecolor=blue,linkcolor=red}

 \newcommand {\etal}{\textit{et al.}}

 \newtheorem{remark}{Remark}
 
 \definecolor{maroon}{rgb}{0.5,0,0}
\begin{document}

\title{Robust Periocular Recognition By Fusing Sparse Representations of Color and Geometry Information}

\author{Juan C. Moreno\footnote{Corresponding author. IT, Department of Computer Science, University of Beira Interior, 6201--001, Covilh\~{a}, Portugal. E-mails: \{jmoreno,gmelfe,hugomcp\}@ubi.pt.}
\and 
V. B. S. Prasath\footnote{Department of Computer Science, University of Missouri-Columbia, MO 65211 USA. E-mail: prasaths@missouri.edu.}
\and
Gil Santos$^*$
\and 
Hugo Proen\c{c}a$^*$}

\date{}
\maketitle

\begin{abstract}

In this paper, we propose a re-weighted elastic net (REN) model for biometric recognition. The new model is applied to data  separated into geometric and color spatial components. The geometric information is extracted using a fast cartoon - texture decomposition model based on a dual formulation of  the total variation norm allowing us to carry information about the overall geometry of images. Color components are defined using linear and nonlinear color spaces, namely the red-green-blue (RGB), chromaticity-brightness (CB) and hue-saturation-value (HSV). Next, according to a Bayesian fusion-scheme, sparse representations for classification purposes are obtained. The scheme is numerically solved using a gradient projection (GP) algorithm. In the empirical validation of the proposed model, we have chosen the \emph{periocular region}, which is an emerging trait known for its robustness against low quality data. Our results were obtained in the publicly available UBIRIS.v2 data set and show consistent improvements in recognition effectiveness when compared to related state-of-the-art techniques.

\end{abstract}

\begin{acronym}
	\acro{DCT}{discrete cosine transform}
	\acro{DWT}{discrete wavelet transform}
	\acro{GEC}{Genetic \& Evolutionary Computing}
	\acro{HOG}{histogram of oriented gradients}
	\acro{LBP}{local binary patters}
	\acro{LoG}{Laplacian of Gaussian}
	\acro{ROI}{region of interest}
	\acro{SIFT}{Scale-Invariant Feature Transform}
	\acro{SURF}{Speed Up Robust Transform}
	\acro{ULBP}{uniform local binary patters}
\end{acronym}

\section{INTRODUCTION}\label{intro}

Biometrics attempts to recognize human beings according to their physical or behavioral features~\cite{JFR07}. In the past, various traits were used for biometric recognition, out of which \emph{iris}  and \emph{face} are the most popular~\cite{PA12IEEE,SJ10,JCL10,PJ10}. Based on the pioneering work of Wright~\etal~\cite{WYGSM09}, the sparse representation theory is emerging as a popular method in the biometrics fields and is considered specially suitable to handle degraded data acquired under uncontrolled acquisition protocols~\cite{PPCR11,SPNChellappa13}.

\subsection{Sparse Representation}\label{sparser}

Model selection in high-dimensional problems has been gaining interest in the statistical signal processing community~\cite{Denoho706, CRT06}. Using convex optimization models, the main problem is recovering a sparse solution $\mathbf{\hat{x}}\in\mathbb{R}^{n}$ of an underdetermined system of the form $\mathbf{y}=A\mathbf{x}^{*}$, given a vector $\mathbf{y}\in\mathbb{R}^{m}$ and a matrix $A\in\mathbb{R}^{m\times n}$. There is a special interest in signal recovery when the number of predictors are much larger than the number of observations (n$\gg$ m). A direct solution to the problem is to select a signal whose measurements are equal to those of $\mathbf{x}^{*}$, with smaller sparsity by  solving a minimization problem based on the $\ell^{0}$-norm:
\begin{equation}\label{l0_intro}
\min_{\mathbf{x}}\|\mathbf{x}\|_{0}\quad\mbox{subj. to $A\mathbf{x}=\mathbf{y}$},
\end{equation} 
($\|\mathbf{x}\|_{0}=\#\{i:\,\,x_{i}\neq 0\}$), being a direct approach to seek the sparsest solution. Problem \eqref{l0_intro} is proved to be NP-hard and difficult to approximate since it involves non-convex minimization~\cite{CTao05}. An alternative method is to relax the problem \eqref{l0_intro} by means of the $\ell^{1}$-norm ($\|\mathbf{x}\|_{1}=\sum_{i=1}^{n}|x_{i}|$). Hence problem \eqref{l0_intro} can be replaced by the following $\ell^{1}$-minimization problem:
\begin{equation*}\label{l1_intro}
\min_{\mathbf{x}}\|\mathbf{x}\|_{1}\quad\mbox{subj. to $A\mathbf{x}=\mathbf{y}$},
\end{equation*}
which can be solved by standard linear programming methods \cite{CDS98}. In practice, signals are rarely exactly sparse, and may often be corrupted by noise. Under noise, the new problem is to reconstruct a sparse signal $\mathbf{y} = A\mathbf{x}^{*}+\mbox{\boldmath$\kappa$}$, where $\mbox{\boldmath$\kappa$}
\in\mathbb{R}^{m}$ is white Gaussian noise with zero mean and variance $\sigma^{2}$. In this case the associated $\ell^{1}$-minimization problem adopts the form:
\begin{equation}\label{l1_intro_1}
\min_{\mathbf{x}}\left\{\tau\|\mathbf{x}\|_{1}+\frac{1}{2}\|\mathbf{y}-A\mathbf{x}\|_{2}^{2}\right\}, 
\end{equation}
where $\tau$ is a nonnegative parameter and $\|\cdot\|_{2}$ denotes the $\ell^{2}$-norm ($\|\mathbf{x}\|_{2}=\left(\sum_{i=1}^{n}x^{2}_{i}\right)^{\frac{1}{2}}$). The convex minimization problem \eqref{l1_intro_1} is known as the least absolute value shrinkage and selection operator (LASSO)~\cite{Tibshirani96}.

Although sparsity of representation seems to be well established by means of the LASSO approach, some limitations were remarked by Hastie \etal~\cite{ZHastie05}. LASSO model tends to select at most $m$ variables before it saturates and in case predictors are highly correlated, LASSO usually selects one variable from a group, ignoring others. In order to overcome these difficulties, Hastie~\etal~\cite{ZHastie05} proposed the elastic net (EN) model as a new regulation technique for outperforming LASSO in terms of prediction accuracy. The elastic net is characterized by the presence of  ridge regression term ($\ell^{2}$-norm) and it is defined by the following convex minimization problem:
\begin{equation}\label{EN1_intro}
\min_{\mathbf{x}}\left\{\tau_{1}\|\mathbf{x}\|_{1} + \tau_{2}\|\mathbf{x}\|_{2}^{2}+\frac{1}{2}\|\mathbf{y}-A\mathbf{x}\|_{2}^{2}\right\},
\end{equation}
where $\tau_{1}$ and $\tau_{2}$ are non-negative parameters. An improvement for the EN model was proposed in \cite{ZZhang09} where a combination of the $\ell^{2}$-penalty and an adaptive version of the $\ell^{1}$-norm have been implemented by considering the minimization problem
\begin{equation}\label{EN2_intro}
\min_{\mathbf{x}}\left\{\tau_{1}\sum_{i=1}^{n}\omega_{i}|x_{i}| + \tau_{2}\|\mathbf{x}\|_{2}^{2}+\frac{1}{2}\|\mathbf{y}-A\mathbf{x}\|_{2}^{2}\right\},
\end{equation}
where the adaptive weights are computed using a solution given by the EN minimization problem \eqref{EN1_intro}. If we let the solution of EN to be $\mathbf{\hat{x}}(EN)$, then the weights are given by the equation $\omega_{i}=1/(|\hat{x}_{i}(EN)|+(1/m))^{\vartheta}$ where $\vartheta$ is a positive constant. A variant of the above model was proposed in \cite{HZhang10} by incorporating the adaptive weight matrix $W$ in the $\ell^{2}$-penalty term:
\begin{equation}\label{EN3_intro}
\min_{\mathbf{x}}\left\{\tau_{1}\sum_{i=1}^{n}\omega_{i}|x_{i}| + \tau_{2}\sum_{i=1}^{n}\omega_{i}^{2}x_{i}^{2}+\frac{1}{2}\|\mathbf{y}-A\mathbf{x}\|_{2}^{2}\right\}.
\end{equation}
In this paper we use a re-weighted elastic net regularization model for periocular recognition application.  
\subsection{Summary of Contributions}\label{summary}

The main contribution of this paper is to propose a re-weighted elastic net (REN) regularization model, that enhances the sparsity of the solutions found. The proposed REN model is a regularization and variable selection method that enjoys  sparsity of representation, particularly when the number of predictors are much larger than the number of observations. The weights are computed such that larger weights will encourage small coordinates by means of the $\ell^{1}$-norm, and smaller weights will encourage large coordinates due to the $\ell^{2}$-norm. Our model differs  from the schemes in~\cite{ZZhang09} and~\cite{HZhang10} (see equations~\eqref{EN2_intro} and~\eqref{EN3_intro} above), since the $\ell^{1}$ and $\ell^{2}$ terms are automatically balanced by weights which are continuously updated using $\omega_{i}=1/(|\hat{x}_{i}|+\gamma)$ with $\gamma$ a positive parameter~\cite{CWBoyd08}. We also provide a concise proof of the existence of a solution for the proposed model as well as its accuracy property. A complete presentation of the numerical implementation of the REN model using a gradient projection (GP) method~\cite{FNW07}, seeking sparse representations along certain gradient directions is described in this paper using a reformulation of the REN model as a quadratic programming (QP) problem. 

\begin{figure}
\centering
\[\begin{array}{c}
\includegraphics[width=.6in,height=.5in]{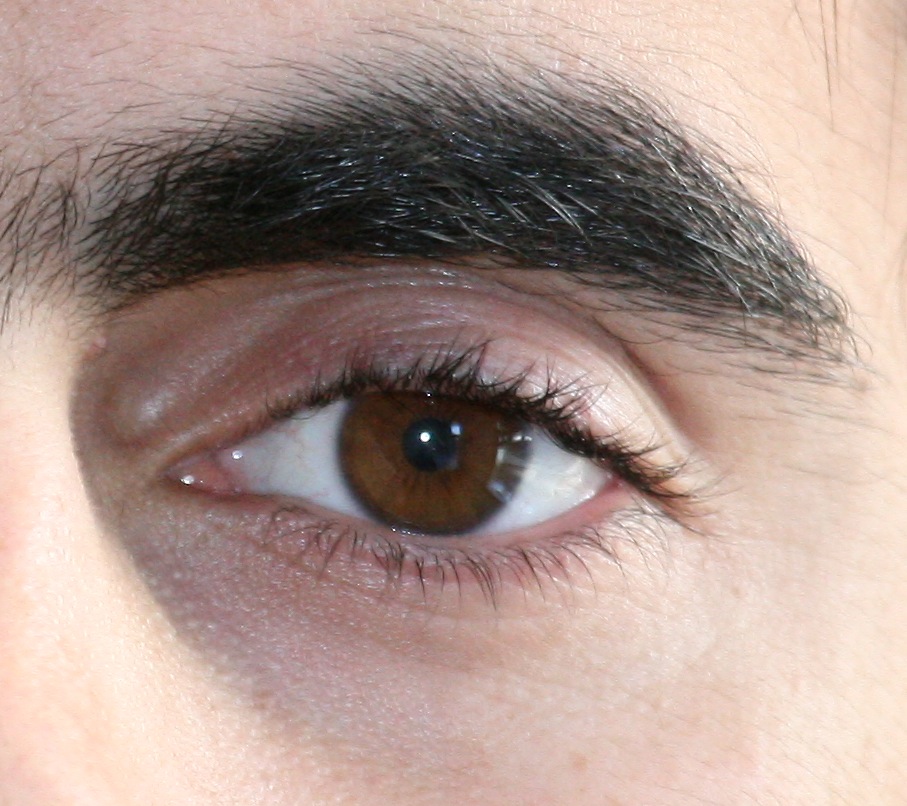} 
\includegraphics[width=.6in,height=.5in]{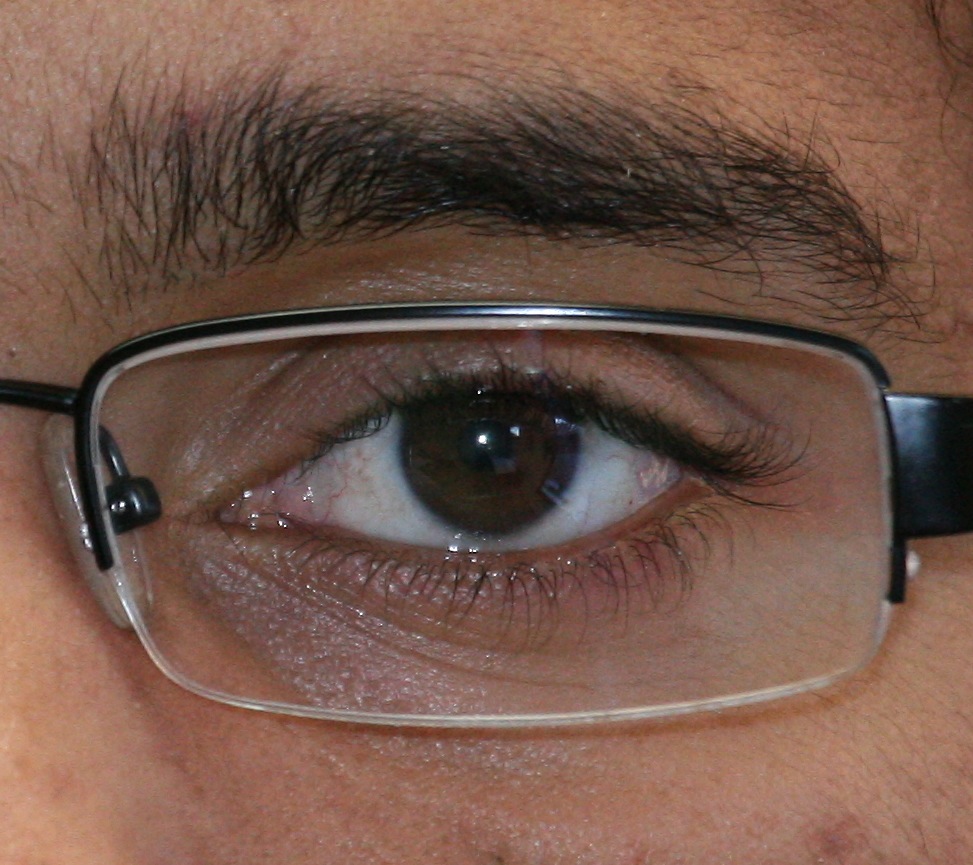} 
\includegraphics[width=.6in,height=.5in]{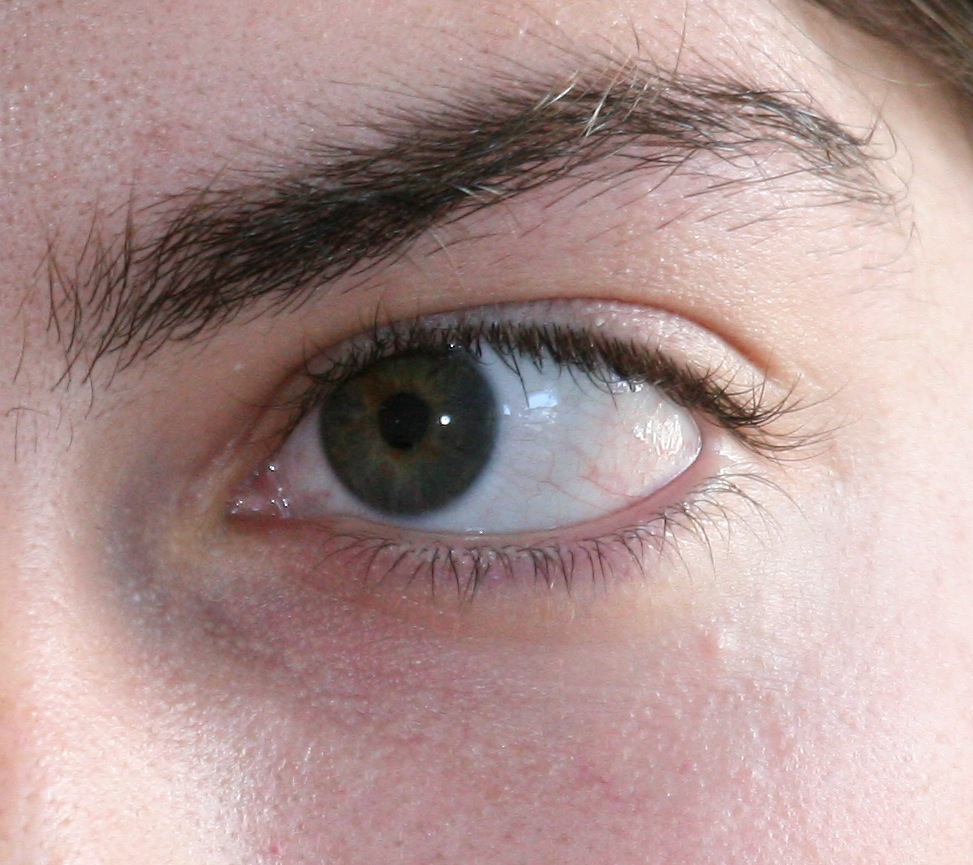}  
\includegraphics[width=.6in,height=.5in]{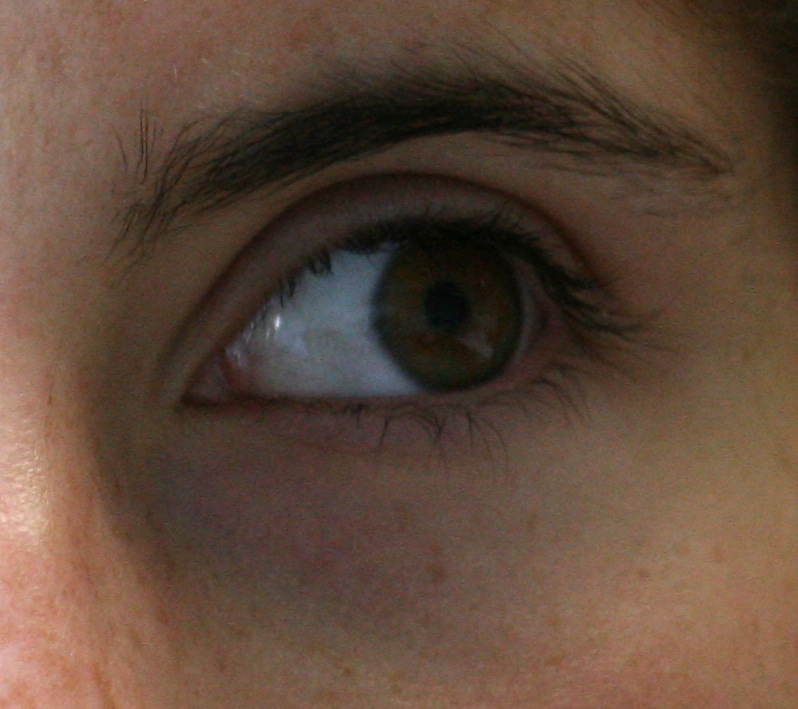}
\includegraphics[width=.6in,height=.5in]{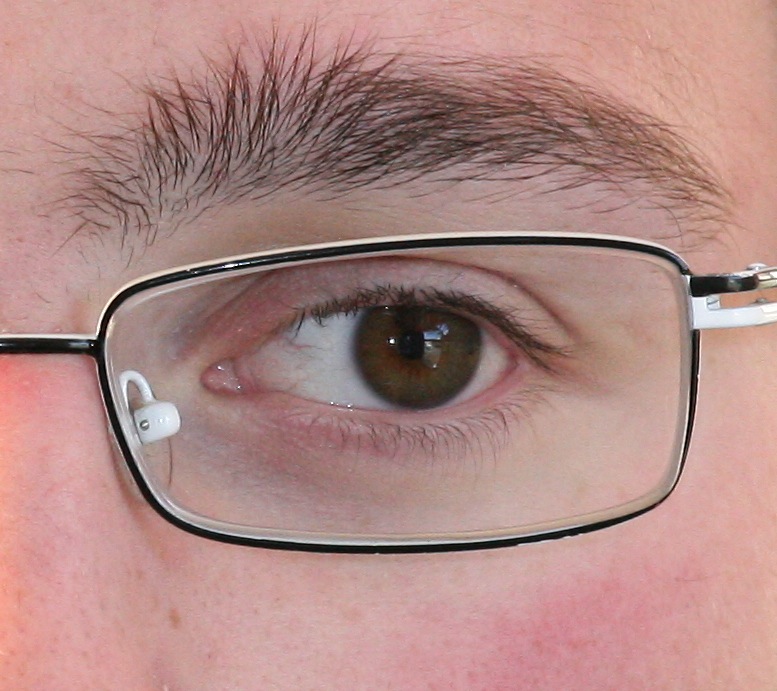}
\includegraphics[width=.6in,height=.5in]{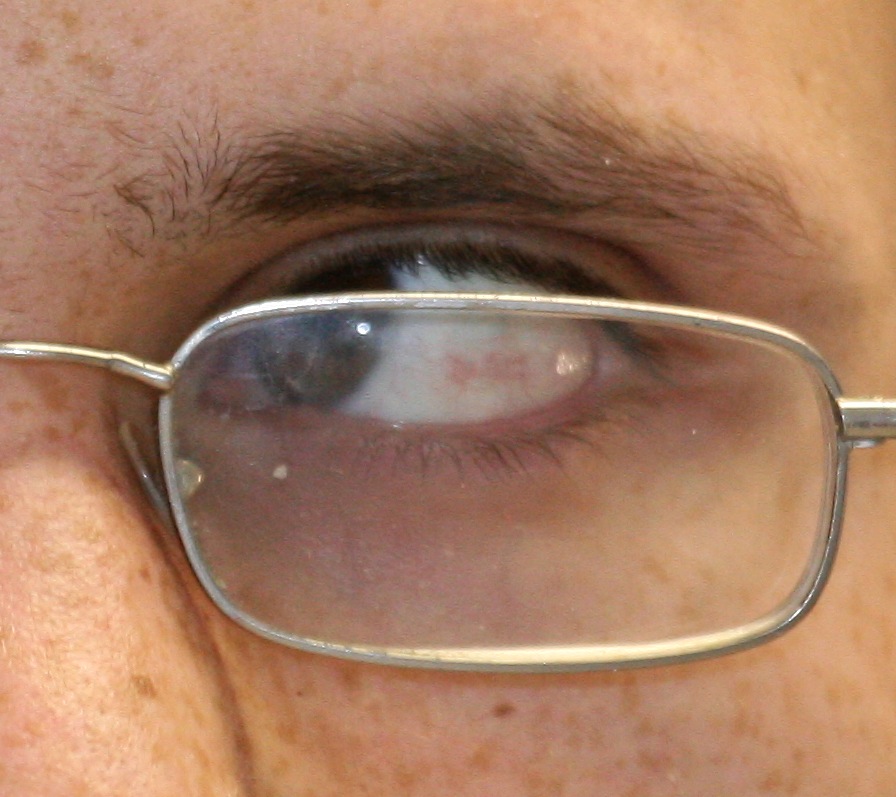} 
\includegraphics[width=.6in,height=.5in]{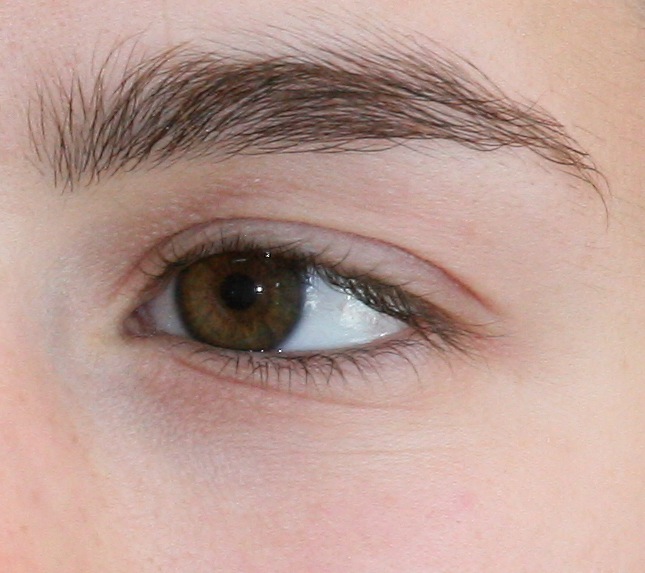} \\
\includegraphics[width=.6in,height=.5in]{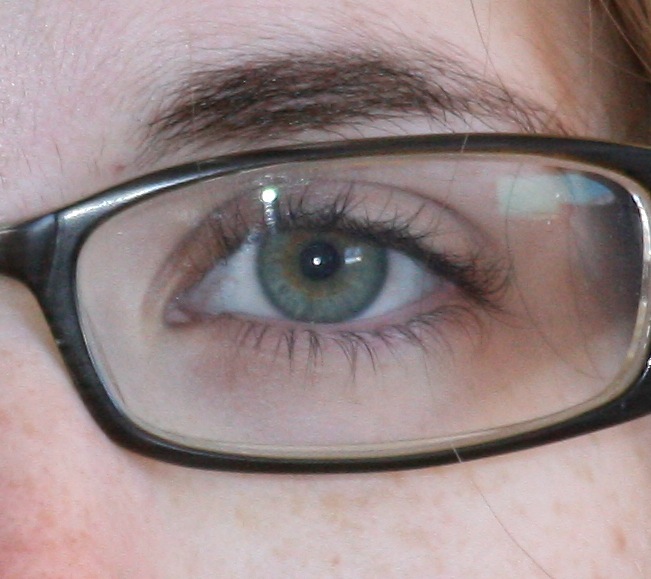}
\includegraphics[width=.6in,height=.5in]{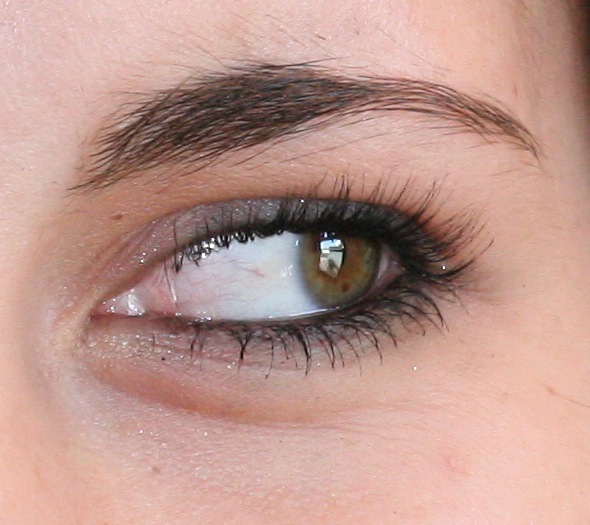}
\includegraphics[width=.6in,height=.5in]{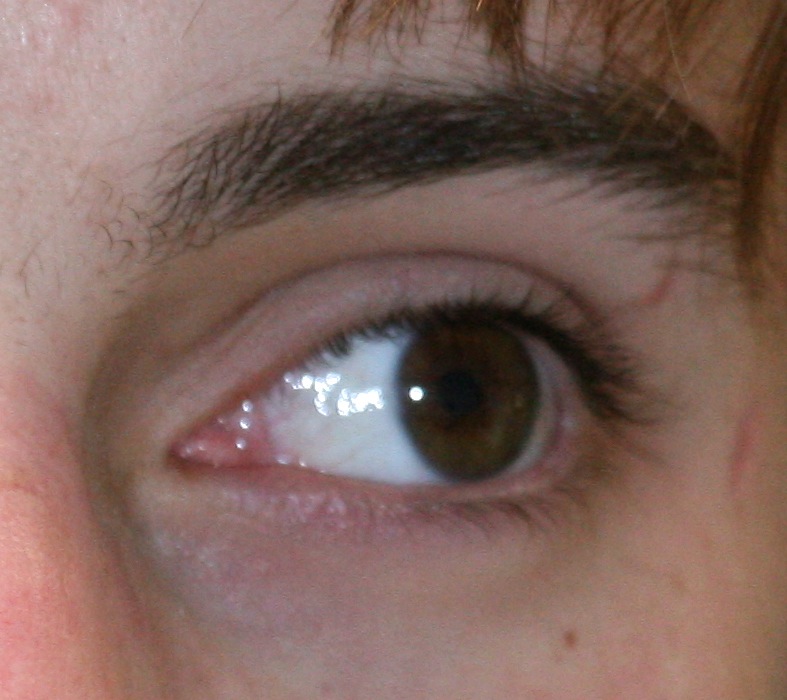}
\includegraphics[width=.6in,height=.5in]{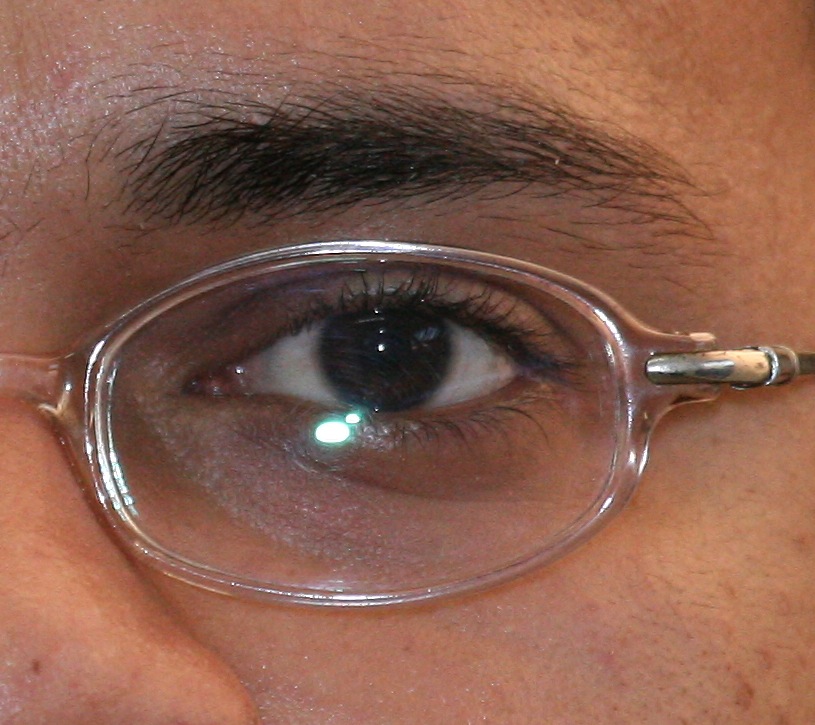} 
\includegraphics[width=.6in,height=.5in]{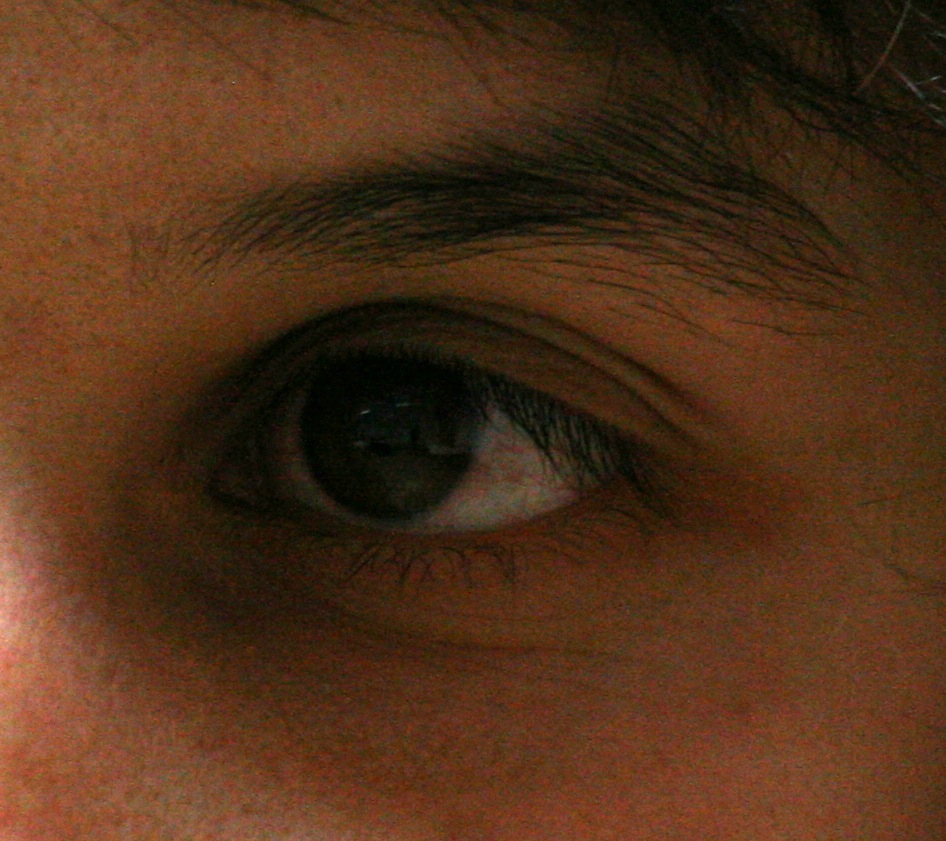}
\includegraphics[width=.6in,height=.5in]{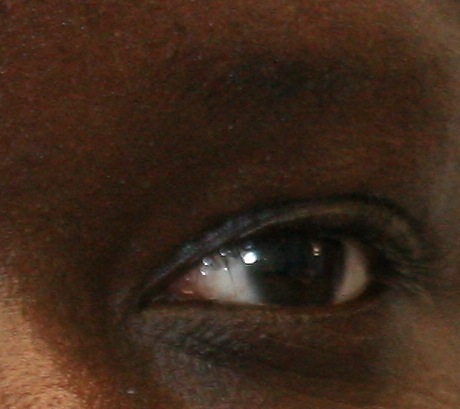}
\includegraphics[width=.6in,height=.5in]{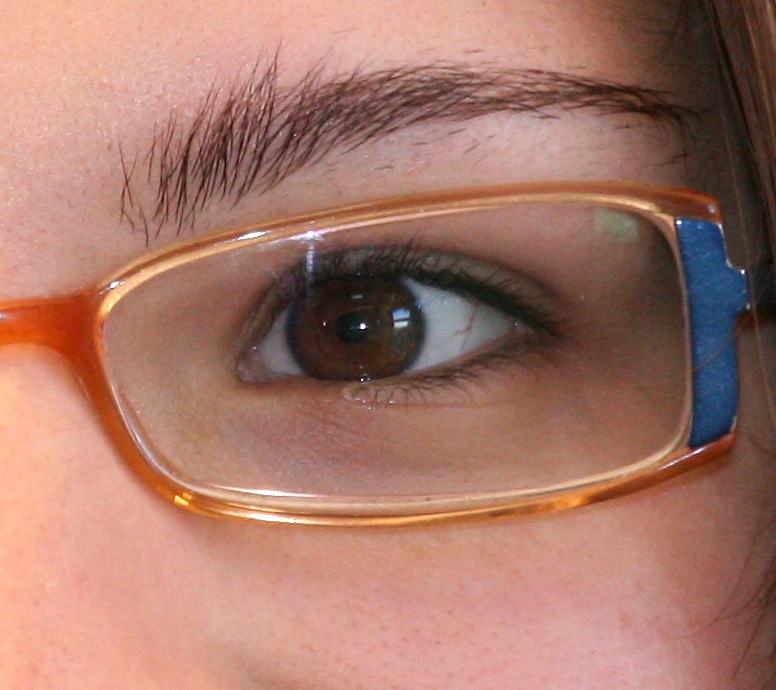}
\\
\end{array}\]
\caption{\scriptsize{Examples of periocular images of different subjects and varying gazes, containing the \emph{corneal}, \emph{eyebrows} and \emph{skin} regions.}}	
\label{fig:Periocular_Dictionaries}
\end{figure}

As a main application of our model, we consider the periocular recognition problem. The periocular region has been regarded as a trade-off between using the entire face or only the iris in biometrics. Periocular region is particularly suitable for recognition under visible wavelength light and uncontrolled acquisition conditions~\cite{PRJ09,WPMJR10,PJRJ11}.  We enhance periocular recognition through the sparsity-seeking property of our REN model over different periocular sectors, which are then fused according to a Bayesian decision based scheme. The main idea is to benefit from the information from each sector, which should contribute in overall recognition robustness. Two different domains are considered for this purpose: (1) \emph{geometry} and (2) \emph{color}. Full geometry information is accessed by decomposing a given image into their cartoon - texture components by means of a dual formulation of the weighted total variation (TV) scheme~\cite{RO92}. For color, a key contribution is the use of nonlinear features such as chromaticity and hue components, which are thought to improve image geometry information according to human perception~\cite{KMarch07}. Our methodology is inspired by two related works: 1) Wright~\etal~\cite{WYGSM09}, which introduced the concept of \emph{sparse representation} for \emph{classification} (SRC) purposes; and 2) Pillai~\etal~\cite{PPCR11}, that used a SRC model for disjoint sectors of the iris and fused results at the score level, according to a confidence score estimated from each sector. Our experiments are carried out in periocular images of the UBIRIS.v2 data set~\cite{ubiris2}: images were acquired at visible wavelengths, from 4 to 8 meters away from the subjects and uncontrolled acquisition conditions. Varying gazes, poses and amounts of occlusions (due to glasses and reflections) are evident in this data set and makes the recognition task harder, see Figure~\ref{fig:Periocular_Dictionaries}. The results obtained using our model allowed us to conclude about consistent increases in performance when compared to the classical SRC model and other important approaches (e.g., Wright~\etal~\cite{WYGSM09} and Pillai~\etal~\cite{PPCR11}). Also, it should be stressed that such increase in performance were obtained without a significant overload in the computational burden of the recognition process.

The reminder of the paper is organized as follows. Section~\ref{rel} summarizes the most relevant in the scope of this work concerning penalized feature selection for sparse representation. The re-weighted elastic net (REN) model is introduced together with statistical motivation ensuring high prediction rates. An algorithm based on gradient projection (GP) for the REN model is also introduced. Section~\ref{Proposed} describes the different geometrical information extracted from periocular images for performing recognition based on cartoon - texture and chromaticity features in a total variation framework. Section~\ref{exper} describes the experimental validation procedure carried out together with remarkable comparisons. Finally, Section~\ref{conc} concludes the paper.


\section{The Reweighted Elastic Net model for Classification Model}\label{rel}
\subsection{The LASSO Model for Recognition}\label{lasso}

We first briefly describe the sparse representation based classification framework which is a precursor to our REN based approach.  Having a set of labeled training samples ($n_{i}$ samples from the i$^{th}$ subject), they are arranged as columns of a matrix $A(i)=[\mathbf{v}_{i,1},\dots,\mathbf{v}_{i,n_{i}}]\in\mathbb{R}^{m\times n_{i}}$. A dictionary results from the concatenation of all samples of all classes:
\begin{align*}
A&=[A(1),\dots,A(k)]=[\mathbf{v}_{1,1},\cdots,\mathbf{v}_{1,n_{1}}|\dots |\mathbf{v}_{k,1},\cdots,\mathbf{v}_{k,n_{k}}]. 
\end{align*}
The key insight is that any probe $\mathbf{y}$ can be expressed as a linear combination of elements of $A$. As the data acquisition process often induces noisy samples, it turns out to be practical  to make use of the LASSO model. In this case it is assumed that the observation model has the form $\mathbf{y} = A\mathbf{x}^{*}+\mbox{\boldmath$\kappa$}$. 

Classification is based on the observation that high values of the coefficients in the solution $\mathbf{\hat{x}}$ are  associated with the columns of $A$ of a single class, corresponding to the identity of the probe. A residual score per class $\mathbbm{1}_{i}:\mathbb{R}^{n}\rightarrow\mathbb{R}^{n}$  is defined: $\mathbf{\hat{x}}\rightarrow\mathbbm{1}_{i}(\mathbf{\hat{x}})$, where $\mathbbm{1}_{i}$ is a indicator function that set the values of all coefficients to 0, except those associated to the i$^{th}$ class. Over this setting, the probe $\mathbf{y}$ is then reconstructed by $\mathbf{\hat{y}}_{i}=A\mathbbm{1}_{i}(\mathbf{\hat{x}})$, and the minimal reconstruction error deemed to correspond to the identity of the probe, between $\mathbf{y}$ and $\mathbf{\hat{y}}_{i}$:
\begin{equation*}\label{PM:classification:e3}
 \text{id}(\mathbf{y})=\arg\min_{i}r_{i}(\mathbf{y}),
 \end{equation*}
 with $r_{i}(\mathbf{y})=\|\mathbf{y}-\mathbf{\hat{y}}_{i}\|_{2}.$
 
In~\cite{WYGSM09} a sparsity concentration index (SCI) is used to accept/reject the response given by the LASSO model. The SCI of a coefficient vector $\mathbf{\hat{x}}\in\mathbb{R}^{n}$ corresponds to:
\begin{eqnarray*}\label{PM:SCI:e1}
SCI(\mathbf{\hat{x}})=\displaystyle\frac{\frac{k\max_{i}\|\mathbbm{1}_{i}(\mathbf{\hat{x}})\|_{1}}{\|\mathbf{\hat{x}}\|_{1}}-1}{k-1}\in [0,1].
\end{eqnarray*}
 If $\text{SCI}(\mathbf{\hat{x}})\approx  1$, the computed signal $\mathbf{\hat{x}}$ is considered to be acceptably represented by samples from a single class. Otherwise, if $\text{SCI}(\mathbf{\hat{x}}) \approx 0$ the sparse coefficients spread evenly across all classes and a reliable identity for that probe cannot be given. 

The recognition model proposed by Pillai~\etal~\cite{PPCR11} obtains separate sparse representations from disjoint regions of an image and fusing them by considering a quality index from each region. Let $L$ be the number of classes with labels $\{c_{i}\}_{i=1}^{L}$. A probe $\mathbf{y}$ is divided into sectors, each one described by the SRC algorithm.  SCI values are obtained over each sector, allowing to reject those with quality bellow a threshold. Let $\{d\}_{i}$ represent the class labels of the retained sectors, and $\mathbb{P}(d_{i}|c)$ be the probability that the $i$-th sector returns a label $d_{i}$, when the true class is $c$:
\begin{equation*}
\mathbb{P}(d_{i}|c)=
\begin{cases}
\frac{t_{1}^{SCI(d_{i})}}{t_{1}^{SCI(d_{i})}+(L-1)t_{2}^{SCI(d_{i})}} & \text{if  $d_{i}=c,$}\\
\frac{t_{2}^{SCI(d_{i})}}{t_{1}^{SCI(d_{i})}+(L-1)t_{2}^{SCI(d_{i})}} & \text{if  $d_{i}\neq c,$}
\end{cases}
\end{equation*}
being $t_{1}$ and $t_{2}$ constants such that $0>t_{1}>t_{2}>1$. According to a maximum a posteriori (MAP) estimate of the class label, the response corresponds to the class having the highest accumulated SCI:
\begin{eqnarray*}\label{FM:eq2}
\tilde{c}=arg\max_{c\in\mathbf{C}} \frac{\sum_{j=1}^{L}SCI(d_{j})\delta(d_{j}=c)}{\sum_{j=1}^{L}SCI(d_{j})}.
\end{eqnarray*}

\subsection{The Re-weighted Elastic Net (REN) Method}\label{ren}

The proposed REN model is a sparsity of representation approach balances the LASSO shrinkage term ($\ell^{1}$-norm) and the strengths of the quadratic regularization ($\ell^{2}$-norm) coefficients by the following minimization problem:
\begin{equation}\label{eq6}
\min_{\mathbf{x}}\left\{\sum_{i=1}^{n}\omega_{i}|x_{i}|+\sum_{i=1}^{n}(1-\omega_{i})^{2}x_{i}^{2}+\frac{1}{2}\|\mathbf{y}-A\mathbf{x}\|_{2}^{2}\right\},
\end{equation} 
where $\omega_{1},\dots,\omega_{n}$ are positive weights taking values in $(0,1)$. The REN-penalty $\sum_{i=1}^{n}\omega_{i}|x_{i}|+\sum_{i=1}^{n}(1-\omega_{i})^{2}x_{i}^{2}$ is strictly convex and it is a compromise between the ridge regression penalty and the LASSO. The convex combination in the REN-penalty term is natural in the sense that both the $\ell^1$ and $\ell^2$ norms are balanced by weights controlling the amount of sparsity versus smoothness expected from the minimization scheme. As in~\cite{CWBoyd08}, the weights are chosen such that they are inversely related to the computed signal according to the equation $\omega_{i}=1/(|\hat{x}_{i}|+\gamma)$ with $\gamma$ a positive parameter. Under this setting, large weights $w_{i}$ will encourage small coordinates with respect to the REN-penalty term, whereas small weights imply big coordinates with respect to the REN-penalty term, respectively. Then, it is seen that the new model combines simultaneously a continuous shrinkage and an automatic variable selection approach. We next consider the existence of solution and the sign recovery property of the REN model. 

\subsection{Existence of Solution}\label{sol}

We state necessary and sufficient conditions for the existence of a solution for the proposed model~\eqref{eq6}. We follow the notations used in~\cite{WWainwright09,JYu10}. In terms of $\ell^{1}$ and $\ell^{2}$ norms, we rewrite the minimization problem in~\eqref{eq6} as,
\begin{equation}\label{MSC:eq4}
\min_{\mathbf{x}}\left\{m\|W\mathbf{x}\|_{1}+\frac{m}{2}\|(1-W)\mathbf{x}\|_{2}^{2}+\frac{1}{2}\|\mathbf{y}-A\mathbf{x}\|_{2}^{2}\right\}.
\end{equation} 
Let us denote by $\mathbf{x}^{*}$ and $\mathbf{\hat{x}}$ the real and estimated solution of~\eqref{MSC:eq4} respectively. Given  $\mathcal{I}=supp(\mathbf{x}^{*})=\{i:\,x^{*}_{i}\neq 0\}$, we define the block-wise form matrix
\begin{equation*}\label{MSC:eq1}
A_{\mathcal{I},\mathcal{I}^{c}}=\frac{1}{m}
\left( \begin{array}{cc}
A^{T}_{\mathcal{I}}A_{\mathcal{I}} & A^{T}_{\mathcal{I}}A_{\mathcal{I}^{c}}   \\
\\
A^{T}_{\mathcal{I}^{c}}A_{\mathcal{I}}  & A^{T}_{\mathcal{I}^{c}}A_{\mathcal{I}^{c}}   \end{array} \right),
\end{equation*}
where $A_{\mathcal{I}}$ ($A_{\mathcal{I}^{c}}$) is a $m\times \#\mathcal{I}$ ($m\times \#\mathcal{I}^{c}$) matrix formed by concatenating the columns $\{A_{i}:\,i\in \mathcal{I}\}$ ($\{A_{i}:\,i\in \mathcal{I}^{c}\}$) and $A^{T}_{\mathcal{I}}A_{\mathcal{I}}$ is assumed to be invertible. 

First we assume that there exist $\mathbf{\hat{x}}\in\mathbb{R}^{n}$ satisfying \eqref{MSC:eq4} and 
$sign(\hat{\mathbf{x}})=sign(\mathbf{x}^{*})$.
Lets define $\mathbf{b}=W_{\mathcal{I}}sign(\mathbf{x}_{\mathcal{I}}^{*})$ together with the set, 
\begin{equation*}\label{MSC:eq6}
\mathcal{D}=\left\{\mathbf{d}\in\mathbb{R}^{n}:\,\begin{cases} d_{i}=b_{i}, &\mbox{for } \hat{x}_{i}\neq 0 \\
|d_{i}|\leq w_{i}, & \mbox{otherwise}\end{cases}\,\,\right\}.
\end{equation*} 
From the Kauush-Kuhn-Tucker (KKT) conditions we obtain
\begin{equation*}
\begin{cases} 
A_{i}^{T}(\mathbf{y}-A\hat{\mathbf{x}})-m(1-w_{i})^{2}\hat{x}_{i}=mw_{i}sign(x^{*}_{i}), &\mbox{if } \hat{x}_{i}\neq 0 \\
\left|A_{i}^{T}\left(\mathbf{y}-A\hat{\mathbf{x}}\right)\right|\leq mw_{i}, & \mbox{otherwise} 
\end{cases}
\end{equation*}
which can be rewritten as,
\begin{equation}\label{MSC:eq8}
A^{T}A(\hat{\mathbf{x}}-\mathbf{x}^{*})-\frac{1}{m}A^{T}\mbox{\boldmath$\kappa$}
+(1-W)^{2}\mathbf{\hat{x}}+\mathbf{d}=0,
\end{equation}
for some $\mathbf{d}\in \mathcal{D}$ and by substituting  the equality $\mathbf{y}=A\mathbf{x}^{*}+\mbox{\boldmath$\kappa$}
$. From the above Eqn.~\eqref{MSC:eq8} the following two equations arise: 
\begin{eqnarray}
A^{T}_{\mathcal{I}}A_{\mathcal{I}}(\hat{\mathbf{x}}_{\mathcal{I}}-\mathbf{x}^{*})-\frac{A_{\mathcal{I}^{c}}\mbox{\boldmath$\kappa$}
}{m}+(1-W)^{2}\hat{\mathbf{x}}_{\mathcal{I}}&=&-\mathbf{b}, \label{MSC:eq9}\\
A^{T}_{\mathcal{I}^{c}}A_{\mathcal{I}}(\hat{\mathbf{x}}_{\mathcal{I}}-\mathbf{x}^{*})-\frac{A^{T}_{\mathcal{I}^{c}}\mbox{\boldmath$\kappa$}
}{m}&=&-\mathbf{d}_{\mathcal{I}^{c}}. \label{MSC:eq10} 
\end{eqnarray}
Solving for $\mathbf{x}_{\mathcal{I}}$ in \eqref{MSC:eq9} and replacing in \eqref{MSC:eq10} to get $\mathbf{b}$ in terms of $\mathbf{x}_{\mathcal{I}}$  leave us with  
\begin{equation} \label{MSC:eq11}\
\mathbf{\hat{x}}_{\mathcal{I}}=\left(A^{T}_{\mathcal{I}}A_{\mathcal{I}}+(1-W)^{2}\right)^{-1}\left(A^{T}_{\mathcal{I}}A_{\mathcal{I}}\mathbf{x}^{*}_{\mathcal{I}}+\frac{A_{\mathcal{I}}\mbox{\boldmath$\kappa$}}{m}-\mathbf{b}\right),
\end{equation}
\begin{equation}\label{MSC:eq12}
A^{T}_{\mathcal{I}^{c}}A_{\mathcal{I}}\left(\left(A^{T}_{\mathcal{I}}A_{\mathcal{I}}+(1-W)^{2}\right)^{-1}\left(A^{T}_{\mathcal{I}}A_{\mathcal{I}}\mathbf{x}_{\mathcal{I}}^{*}+\frac{A_{\mathcal{I}}^{T}\mbox{\boldmath$\kappa$}
}{m}-\mathbf{b}\right)-\mathbf{x}^{*}_{\mathcal{I}}\right)-\frac{A^{T}_{\mathcal{I}^{c}}\mbox{\boldmath$\kappa$}
}{m}=-\mathbf{b}. 
\end{equation}
From \eqref{MSC:eq11} and \eqref{MSC:eq12}, we finally get the next two equations:
\begin{equation}\label{MSC:eq13}
sign\left(\left(A^{T}_{\mathcal{I}}A_{\mathcal{I}}+(1-W)^{2}\right)^{-1}\left(A^{T}_{\mathcal{I}}A_{\mathcal{I}}\mathbf{x}^{*}_{S}+\frac{A_{\mathcal{I}}^{T}\mbox{\boldmath$\kappa$}}{m}-\mathbf{b}\right)\right)=sign(\mathbf{x}^{*}_{\mathcal{I}})
\end{equation}
and
\begin{equation}\label{MSC:eq14}
\left|A_{i}^{T}A_{\mathcal{I}}\left(\left(A^{T}_{\mathcal{I}}A_{\mathcal{I}}+(1-W)^{2}\right)^{-1}\left(A^{T}_{\mathcal{I}}A_{\mathcal{I}}\mathbf{x}_{\mathcal{I}}^{*}+\frac{A_{\mathcal{I}}^{T}\mbox{\boldmath$\kappa$}
}{m}-\mathbf{b}\right)-\mathbf{x}^{*}_{\mathcal{I}}\right)-\frac{A_{i}^{T}\mbox{\boldmath$\kappa$}
}{m}\right|\leq w_{i},
\end{equation}
for $i\in\mathcal{I}^{c}$.

Now, let us assume that equations~\eqref{MSC:eq13} and~\eqref{MSC:eq14} both hold. It will be proved there exist $\hat{\mathbf{x}}\in\mathbb{R}^{n}$ satisfying  $sing(\hat{\mathbf{x}})=sign(\mathbf{x}^{*})$. Setting $\hat{\mathbf{x}}\in\mathbb{R}^{n}$ satisfying $\hat{\mathbf{x}}_{\mathcal{I}^{c}}=\mathbf{x}^{*}_{\mathcal{I}^{c}}=0$ and
\begin{equation*}
\mathbf{x}_{\mathcal{I}}=\left(A^{T}_{\mathcal{I}}A_{\mathcal{I}}+(1-W)^{2}\right)^{-1}\left(A^{T}_{\mathcal{I}}A_{\mathcal{I}}\mathbf{x}^{*}_{\mathcal{I}}+\frac{A_{\mathcal{I}}^{T}\mbox{\boldmath$\kappa$}
}{m}-\mathbf{b}\right),
\end{equation*}
which guarantees the equality  $sign(\mathbf{\hat{x}}_{\mathcal{I}})=sign(\mathbf{x}^{*}_{\mathcal{I}})$  due to \eqref{MSC:eq13}.
In the same manner, we define $\mathbf{d}\in\mathbb{R}^{n}$ satisfying $\mathbf{d}_{\mathcal{I}}=\mathbf{b}$ and 
\begin{equation*}
\begin{aligned}
\mathbf{d}_{\mathcal{I}^{c}}=-\left(A^{T}_{\mathcal{I}^{c}}A_{\mathcal{I}}\left(\left(A^{T}_{\mathcal{I}}A_{\mathcal{I}}+(1-W)^{2}\right)^{-1}\left(A^{T}_{\mathcal{I}}A_{\mathcal{I}}\mathbf{x}_{\mathcal{I}}^{*}+\frac{A_{\mathcal{I}}^{T}\mbox{\boldmath$\kappa$}}{m}-\mathbf{b}\right)-\mathbf{x}^{*}_{\mathcal{I}}\right)-
\frac{A_{\mathcal{I}^{c}}^{T}\mbox{\boldmath$\kappa$}}{m}\right),
\end{aligned}
\end{equation*}
implying from \eqref{MSC:eq14} the inequality $|d_{i}|\leq w_{i}$ for $i\in\mathcal{I}^{c}$ and therefore $\mathbf{d}\in\mathcal{D}$. From previous, we have found a point a point $\mathbf{\hat{x}}\in\mathbb{R}^{n}$ and $\mathbf{d}\in\mathcal{D}$ satisfying \eqref{MSC:eq9} and \eqref{MSC:eq10} respectively or equivalently \eqref{MSC:eq8}. Moreover, we also have the equality $sign(\mathbf{\hat{x}})=sign(\mathbf{x}^{*})$. Under these assertions we can prove the sign recovery property of our model as illustrated next.

\subsection{Sign Recovery Property}\label{sign}

Under some regularity conditions on the proposed REN model, we intend to give an estimation for which the event $sign(\mathbf{\hat{x}})=sign(\mathbf{x}^{*})$ is true. Following similar notations in~\cite{ZZhang09,HZou06}, we intend to prove that our model enjoys the following probabilistic property:
\begin{equation}\label{SRP:eq55}
Pr\left(\min_{i\in\mathcal{I}}
\left|\hat{x}_{i}\right|>0\right)\rightarrow 1.
\end{equation}
For theoretical analysis purposes, the problem~\eqref{eq6} is written as
\begin{equation*}\label{MSC:eq555}
\min_{\mathbf{x}}\left\{\|W\mathbf{x}\|_{1}+\|(1-W)\mathbf{x}\|_{2}^{2}+\|\mathbf{y}-A\mathbf{x}\|_{2}^{2}\right\}.
\end{equation*}  
The following regularity conditions are also assumed:
\begin{itemize}

\item[1.] Denoting with $\Lambda_{\min}(S)$ and $\Lambda_{\max}(S)$ the minimum and maximum eigenvalues of a symmetric matrix $S$, we assume the following inequalities hold:
\begin{equation*}\label{SRP:eq6}
\theta_{1}\leq\Lambda_{\min}\left(\frac{1}{m}A^{T}A\right)\leq\Lambda_{\max}\left(\frac{1}{m}A^{T}A\right)\leq\theta_{2},
\end{equation*} 
where $\theta_{1}$ and $\theta_{2}$ are two positive constants.

\item[2.] $\displaystyle\lim_{m\rightarrow\infty}\frac{\log(n)}{\log(m)}=\nu$ for some $0\leq\nu<1$
\item[3.] $\displaystyle\lim\limits_{m\rightarrow\infty}\sqrt{\frac{m}{n}}\frac{1}{\max_{i\in\mathcal{I}}w_{i}}=\infty$.

\end{itemize}
Let
\begin{equation}\label{SRP:eq7}
\mathbf{\tilde{x}}=arg\min_{\mathbf{x}}\left\{\left\|\mathbf{y}-A\mathbf{x}\right\|_{2}^{2}+\left\|(1-W)\mathbf{x}\right\|_{2}^{2}\right\}.
\end{equation}
By using the definitions of $\mathbf{\hat{x}}$ and $\mathbf{\tilde{x}}$, the next two inequalities arise
\begin{equation}\label{SRP:eq8}
\left\|\mathbf{y}-A\mathbf{\hat{x}}\right\|_{2}^{2}+\left\|\left(1-W\right)\mathbf{\hat{x}}\right\|_{2}^{2}\geq
\left\|\mathbf{y}-A\mathbf{\tilde{x}}\right\|_{2}^{2}+\left\|\left(1-W\right)\mathbf{\tilde{x}}\right\|_{2}^{2}
\end{equation}
and
\begin{equation}\label{SRP:eq9}
\begin{aligned}
&\left\|\mathbf{y}-A\mathbf{\tilde{x}}\right\|_{2}^{2}+\left\|\left(1-W\right)\mathbf{\tilde{x}}\right\|_{2}^{2}
+\sum_{i=1}^{n}w_{i}|\tilde{x}_{i}|
\geq
\left\|\mathbf{y}-A\mathbf{\hat{x}}\right\|_{2}^{2}+\left\|\left(1-W\right)\mathbf{\hat{x}}\right\|_{2}^{2}
+\sum_{i=1}^{n}w_{i}|\hat{x}_{i}|.
\end{aligned}
\end{equation}
The combination of equations \eqref{SRP:eq8} and \eqref{SRP:eq9} give 
\begin{equation}\label{SRP:eq10}
\begin{aligned}
\sum_{i=1}^{n}w_{i}(|\tilde{x}_{i}|-|\hat{x}_{i}|)&\geq
\left\|\mathbf{y}-A\mathbf{\hat{x}}\right\|_{2}^{2}+\left\|(1-W)\mathbf{\hat{x}}\right\|_{2}^{2}
-\left\|\mathbf{y}-A\mathbf{\tilde{x}}\right\|_{2}^{2}-\left\|(1-W)\mathbf{\tilde{x}}\right\|_{2}^{2}\\
&=\left(\mathbf{\hat{x}}-\mathbf{\tilde{x}}\right)^{T}
\left(A^{T}A+(1-W)^{2}\right)
\left(\mathbf{\hat{x}}-\mathbf{\tilde{x}}\right)\\
\end{aligned}
\end{equation}
On the other hand
\begin{equation}\label{SRP:eq11}
\begin{aligned}
\sum_{i=1}^{n}w_{i}\left(\left|\tilde{x}_{i}\right|-\left|\hat{x}_{i}\right|\right)&\leq
\sum_{i=1}^{n}w_{i}\left|\tilde{x}_{i}-\hat{x}_{i}\right|\leq\sqrt{\sum_{i=1}^{n}w_{i}^{2}}\left\|\mathbf{\tilde{x}}-\mathbf{\hat{x}}\right\|_{2}\\
\end{aligned}
\end{equation}
By combining equations \eqref{SRP:eq10} and \eqref{SRP:eq11} we get
\begin{equation*}
\begin{aligned}
\Lambda_{min}\left(\left(A^{T}A\right)+\left(1-W\right)^{2}\right)\left\|\mathbf{\hat{x}}-\mathbf{\tilde{x}}\right\|_{2}^{2}&
\leq\left(\mathbf{\hat{x}}-\mathbf{\tilde{x}}\right)^{T}
\left(A^{T}A+(1-W)^{2}\right)
\left(\mathbf{\hat{x}}-\mathbf{\tilde{x}}\right)\\
&\leq \sqrt{\sum_{i=1}^{n}w_{i}^{2}}\left\|\mathbf{\tilde{x}}-\mathbf{\hat{x}}\right\|_{2}  
\end{aligned}
\end{equation*}
which together with the identity
\begin{equation*}
0\leq\theta_{1}\leq\Lambda_{min}\left(A^{T}A\right)\leq\Lambda_{min}\left(\left(A^{T}A\right)+\left(1-W\right)^{2}\right)
\end{equation*}
allow us to prove
\begin{equation}\label{SRP:eq12}
\left\|\mathbf{\hat{x}}-\mathbf{\tilde{x}}\right\|_{2}\leq\frac{\sqrt{\sum_{i=1}^{n}w_{i}^{2}}}{\Lambda_{min}\left(A^{T}A\right)},
\end{equation}
Let us notice that
\begin{equation}\label{SRP:eq13}
\begin{aligned}
E\left(\left\|\mathbf{\tilde{x}}-\mathbf{x}^{*}\right\|_{2}^{2}\right)
&=E\left(-\left(A^{T}A+\left(1-W\right)^{2}\right)^{-1}\left(1-W\right)^{2}\mathbf{x}^{*}
+\left(A^{T}A+\left(1-W\right)^{2}\right)^{-1}A^{T}\mbox{\boldmath$\kappa$}\right)\\
&\leq2\frac{\left\|(1-W)\mathbf{x}^{*}\right\|_{2}^{2}+n\Lambda_{\max}\left(A^{T}A\right)\sigma^{2}}{\Lambda_{\min}\left(A^{T}A\right)}
\end{aligned}
\end{equation}
From equations \eqref{SRP:eq12} and \eqref{SRP:eq13} we conclude that
\begin{equation}\label{SRP:eq14}
\begin{aligned}
E\left(\left\|\mathbf{\hat{x}}-\mathbf{x}^{*}\right\|_{2}^{2}\right)
&\leq2\left(E\left(\left\|\mathbf{\tilde{x}}-\mathbf{x}^{*}\right\|_{2}^{2}\right)
-E\left(\left\|\mathbf{\hat{x}}-\mathbf{x}^{*}\right\|_{2}^{2}\right)\right)\\
&\leq4\frac
{\left\|\left(1-W\right)\mathbf{x}^{*}\right\|_{2}^{2}+n\Lambda_{\max}(A^{T}A)\sigma^{2}+E\left(\sum_{i=1}^{n}w_{i}^{2}\right)}
{\Lambda_{\min}\left(A^{T}A\right)}.
\end{aligned}
\end{equation}

Let $\eta=\min_{i\in\mathcal{I}}|x_{i}^{*}|$ and $\hat{\eta}=\max_{i\in\mathcal{I}}w_{i}$. Because of \eqref{SRP:eq12}, 
\begin{equation*}\label{SRP:eq15}
\left\|\mathbf{\hat{x}}_{\mathcal{I}}-\mathbf{\tilde{x}}_{\mathcal{I}}\right\|_{2}^{2}\leq\frac{\sqrt{n}\hat{\eta}}{\theta_{1}m}.
\end{equation*}
Then
\begin{equation}\label{SRP:eq16}
\min_{i\in\mathcal{I}}|x^{*}_{i}|>\min_{i\in\mathcal{I}}|\tilde{x}_{i}|-\frac{\sqrt{n}\hat{\eta}}{\theta_{1}m}>\min_{i\in\mathcal{I}}|\hat{x}_{i}|-
\left\|\mathbf{\tilde{x}}_{\mathcal{I}}-\mathbf{x}^{*}_{\mathcal{I}}\right\|_{2}
-\frac{\sqrt{n}\hat{\eta}}{\theta_{1}m}.
\end{equation}
Now, we notice that
\begin{equation*}\label{SRP:eq17}
\displaystyle\frac{\sqrt{n}\hat{\eta}}{\theta_{1}m}=O\left(\frac{1}{\sqrt{n}}\right)
\left(\sqrt{\frac{n}{m}}\eta^{-1}\right)\left(\hat{\eta}\eta\right).
\end{equation*}
Since
\begin{equation*}\label{SRP:eq18}
\begin{aligned}
E\left(\left(\hat{\eta}\eta\right)^{2}\right)
&\leq 2\eta^{2}+2\eta^{2}E\left(\left(\hat{\eta}-\eta\right)^{2}\right)
\leq 2\eta^{2}+2\eta^{2}E\left(\left\|\mathbf{\hat{x}}-\mathbf{x}^{*}\right\|^{2}\right)\\
&\leq 2 \eta^{2}+8\eta^{2}\frac{\left\|\left(1-W\right)^{2}\mathbf{x}^{*}\right\|_{2}^{2}+\theta_{2}nm\sigma^{2}+E\left(\sum_{i=1}^{n}w_{i}^{2}\right)}{\theta_{1}m}
\end{aligned}
\end{equation*}
and $\eta^{2}m/n\rightarrow\infty$ as long as $m\rightarrow\infty$, it follows that 
\begin{equation}\label{SRP:eq19}
\displaystyle\frac{\sqrt{n}\hat{\eta}^{-1}}{\theta_{1}m}=o\left(\frac{1}{\sqrt{n}}\right)O_{Pr}(1).
\end{equation}
By using \eqref{SRP:eq14}, we derive
\begin{equation}\label{SRP:eq20}
E\left(\left\|\mathbf{\hat{\hat{x}}}_{\mathcal{I}}-\mathbf{x}^{*}_{\mathcal{I}}\right\|_{2}^{2}\right)
\leq4\frac{\|\left(1-W\right)^{2}\mathbf{x}^{*}\|_{2}+\theta_{2}nm\sigma^{2}}{(\theta_{1}m)^{2}}=\sqrt{\frac{n}{m}}O_{Pr}(1).
\end{equation}
Substituting \eqref{SRP:eq19} and \eqref{SRP:eq20} in  \eqref{SRP:eq16} allow us to conclude that
\begin{equation*}
\min_{i\in\mathcal{I}}|x^{*}_{i}|>\eta-\sqrt{\frac{n}{m}}O_{Pr}(1)-o\left(\frac{1}{\sqrt{n}}\right)O_{Pr}(1).
\end{equation*} 
Then \eqref{SRP:eq55} holds.
 
\begin{remark}
There is special interest in applying the REN model in the case the data satisfies the condition $n\gg m$. For the LASSO model it was suggested in~\cite{CTao07} to make use of the Dantzig selector which can achieve the ideal estimation up to a $log(n)$ factor. In~\cite{FLv08} a performing of the Dantzig selector called the Sure Independence Screening (SIS) was introduced in order to  reduce the ultra-high dimensionality. We remark that the SIS technique can be combined with the REN model~\eqref{eq6} for dealing the case $n\gg m$. Then previous computations can be still applied to reach the sign recovery property.  
\end{remark}
Next we describe an algorithm for the REN model allowing us to directly deal with the case $n\gg m$.  It turns out that our REN model can be expressed as a quadratic program (QP), thus allowing us to apply a gradient projection approach to perform the sparse reconstruction.
\begin{figure}
\centering
\[\begin{array}{c}
\mbox{\scriptsize{(a) Original Signal}}\\
\includegraphics[width=10cm,height=0.8cm]{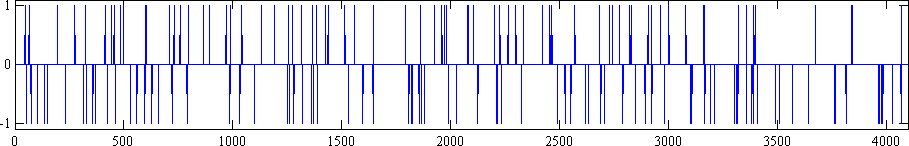}\\ 
\mbox{\scriptsize{(b) Reweighted EN Model (Proposed) (MSE = 4.499e-05)}}\\
\includegraphics[width=10cm,height=0.8cm]{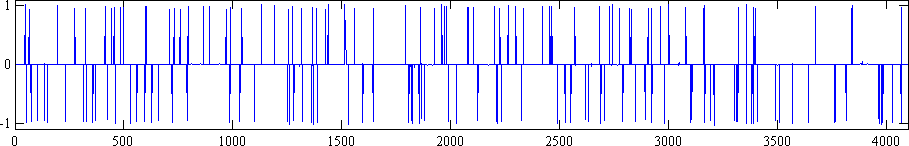}\\
\mbox{\scriptsize{(c) Adaptive EN Model \cite{ZZhang09} ( MSE = 5.194e-05)}}\\
\includegraphics[width=10cm,height=0.8cm]{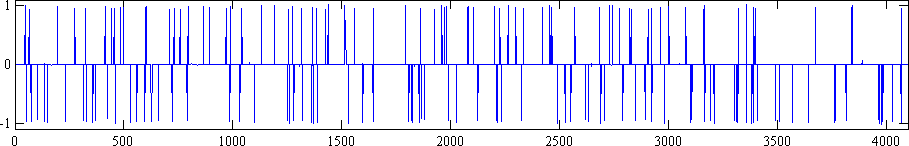}\\
\mbox{\scriptsize{(d) Adaptive EN Model \cite{HZhang10} (MSE = 4.791e-05)}}\\
\includegraphics[width=10cm,height=0.8cm]{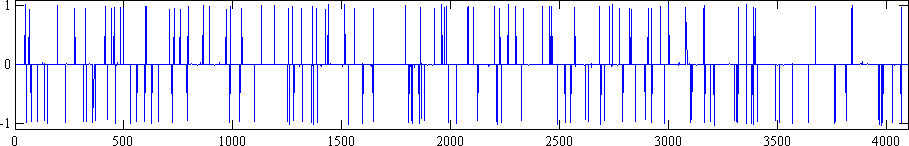}\\
\mbox{\scriptsize{(e) LASSO Model (MSE = 1.445e-04)}}\\ 
\includegraphics[width=10cm,height=0.8cm]{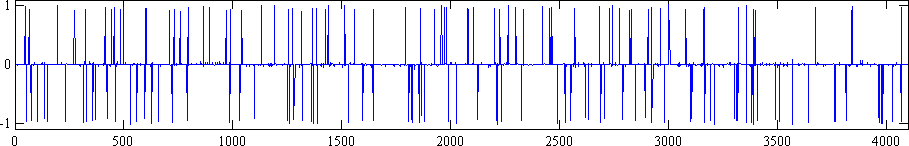}\\
\end{array}\]
\caption{\scriptsize{Sparse signal reconstruction with EN and LASSO models. (a) Sparse signal of Length $n=4096$ with $k=1024$ observations. (b)-(e) Response signals computed with the proposed reweighted elastic net, \cite{ZZhang09}, \cite{HZhang10} and LASSO, respectively.}} 	
\label{fig:Reweigthed_EN}
\end{figure}

\subsection{Numerical Implementation}\label{numer}

The algorithm that alternates between the computed signal and redefining the weights is as follows:
\begin{enumerate}
\item Choose initial weights $w_{i}=1/2$, $i=1,\dots, n$.
\item Find the solution $\mathbf{\hat{x}}$ of the problem
\begin{equation}\label{EN:eq1}
\min_{\mathbf{x}}\|W\mathbf{x}\|_{1}+\|(1-W)\mathbf{x}\|_{2}^{2}+\frac{1}{2}\|\mathbf{y}-A\mathbf{x}\|_{2}^{2},
\end{equation}
\item Update the weights: for each $i=1,\cdots,n$,
\begin{equation*}\label{EN:eq2}
w_{i}=\displaystyle\frac{1}{\left |\hat{x}_{i}\right|+\gamma},
\end{equation*}
where $\gamma$ is a positive stability parameter.
\item Terminate on convergence or when a specific number of iterations is reached. Otherwise, go to step 2.
\end{enumerate}

Note that our REN problem in~\eqref{EN:eq1} can also be expressed as a quadratic program~\cite{Fuchs}, by splitting the variable $\mathbf{x}$ into its positive and negative parts. That is, $\mathbf{x}=\mathbf{x}_{+}-\mathbf{x}_{-},$ where $\mathbf{x}_{+}$ and $\mathbf{x}_{-}$ are the vectors that collect the positive and negative coefficients of $\mathbf{x}$, respectively. 
Then, we handle the minimization problem,
\begin{equation}\label{Equation11}
\min_{\mathbf{z}}\left\{Q(\mathbf{z})= \mathbf{c}^{T}\mathbf{z}+\mathbf{z}^{T}B\mathbf{z}\right\},
\end{equation}
where $\mathbf{z}=[\mathbf{x}_{+},\mathbf{x}_{-}]^{T}$, $\mathbf{w}_{n}=[\omega_{1},\dots,\omega_{n}]^{T}$, $\mathbf{c}=\mathbf{w}_{2n}+[-A^{T}\mathbf{y};A^{T}\mathbf{y}]^{T}$ and $B=\frac{1}{2}B_{1}+B_{2}$ with
\begin{equation*}\label{Equation12}
B_{1}= \left( \begin{array}{cc}
 A^{T}A&-A^{T}A \\
-A^{T}A & A^{T}A 
 \end{array} \right), \quad \quad
 B_{2}= \left( \begin{array}{cc}
 (1-W)^{2}&-(1-W)^{2} \\
-(1-W)^{2} & (1-W)^{2} 
 \end{array} \right).
\end{equation*} 
The minimization problem~\eqref{Equation11} can then be solve using the Barzilai-Borwein Gradient Projection Algorithm~\cite{SZZanni03}. Under this approach the iterative equation is given by, 
\begin{equation*}\label{Equation16}
\mathbf{z}^{(k+1)}=\mathbf{z}^{(k)}-\zeta^{(k)}\nu^{(k)},
\end{equation*}
where $\zeta^{(k)}$  is the step size computed as
\begin{equation*}
\zeta^{(k)}=\left(\mathbf{z}^{(k)}-\alpha^{(k)}\nabla Q\left(\mathbf{z}^{(k)}\right)\right)_{+}-\mathbf{z}^{(k)},
\end{equation*} 
with 
\begin{equation*}
\displaystyle
\alpha^{(k+1)}=
\begin{cases}
mid\left\{\alpha_{min},\displaystyle\frac{\left\|\zeta^{(k)}\right\|^{2}}{\left(\zeta^{(k)}\right)^{T}B\zeta^{(k)}},\alpha_{max}\right\}, & \mbox{if $\left(\zeta^{(k)}\right)^{T}B\zeta^{(k)}\neq 0$}\\
\alpha_{max}, & \mbox{otherwise.}
\end{cases}
\end{equation*}
The operator $mid$ is the define as the middle value of three scalar arguments and $\alpha_{min}$ and $\alpha_{max}$ are two given parameters. The parameter $\nu$ take the form
\begin{equation*}
\nu^{(k)}=
\begin{cases}
mid\left\{0,\displaystyle\frac{\left(\zeta^{(k)}\right)^{T}\nabla Q\left(\mathbf{z}^{(k)}\right)}{\left(\zeta^{(k)}\right)^{T}B\zeta^{(k)}},1\right\} , & \mbox{if $\left(\zeta^{(k)}\right)^{T}B\zeta^{(k)}\neq 0$,}\\
1, & \mbox{otherwise.}
\end{cases}
\end{equation*}

The performance of the REN minimization along with comparisons is shown is Figure \ref{fig:Reweigthed_EN} for a sparse signal. We want to reconstruct a length-$n$ sparse signal (in the canonical basis) from $m$ observations, with $m\ll n$. The matrix $A_{m\times n}$ is build with independent samples of a standard Gaussian distribution and by ortho-normalizing the rows, while the original signal $\mathbf{x}^{*}$ contains 160 randomly placed $\pm spikes$ and the observation is defined as $\mathbf{y}=A\mathbf{x}^{*}+\mbox{\boldmath$\kappa$}$ with $\mbox{\boldmath$\kappa$}$ a Gaussian noise of variance $\sigma^{2}=10^{-4}$. The reconstruction of the original signal over the REN minimization problem produces a much lower mean squared error (MSE = $(1/n) \|\mathbf{\hat{x}}-\mathbf{x}^{*}\|$ with $\hat{\mathbf{x}}$ been an estimate of $\mathbf{x}^{*}$) equal to $4.499\times 10^{-05}$, while the MSE given by the adaptive elastic model proposed in \cite{HZhang10}, \cite{ZZhang09}  and LASSO are $5.194\times 10^{-05}$, $4.791\times 10^{-05}$ and $1.445\times 10^{-04}$ respectively. Therefore, the proposed REN approach does an excellent job at locating the spikes.

\begin{figure}
\centering
\[\begin{array}{c}
\includegraphics[width=.7in,height=.6in]{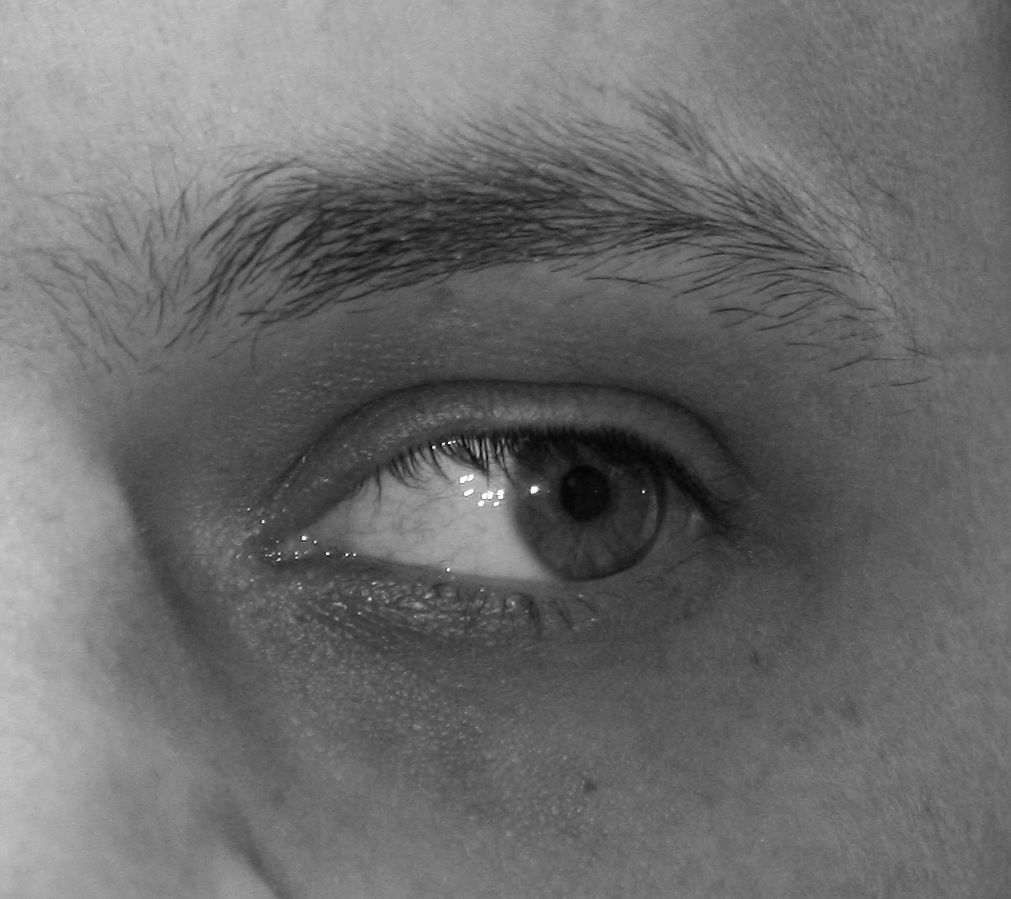}  \includegraphics[width=.7in,height=.6in]{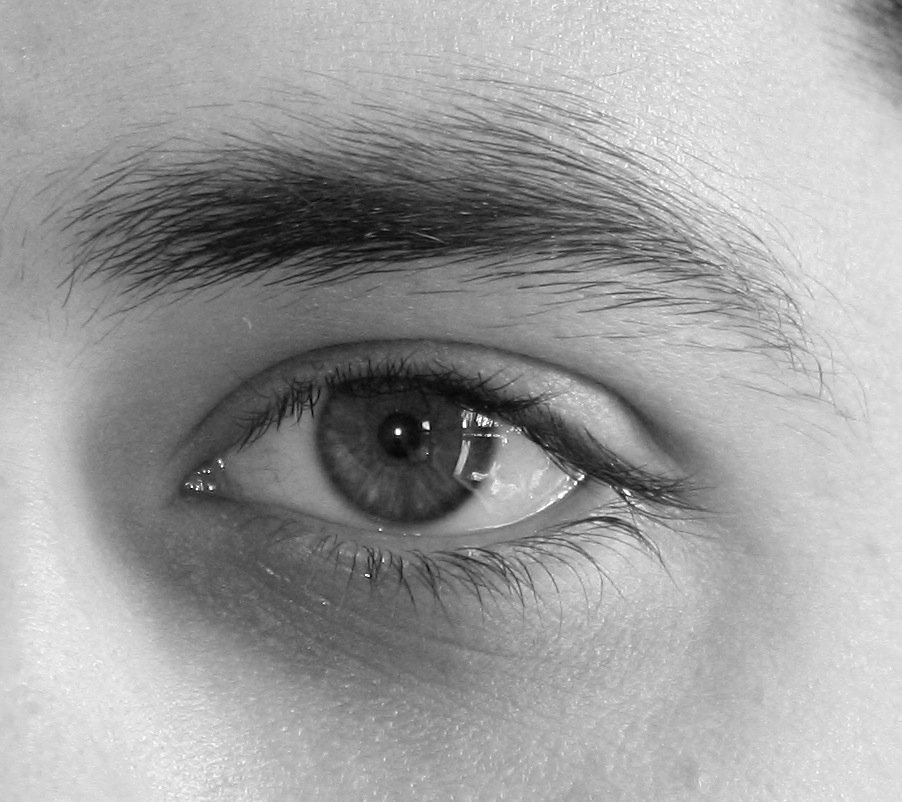}  \includegraphics[width=.7in,height=.6in]{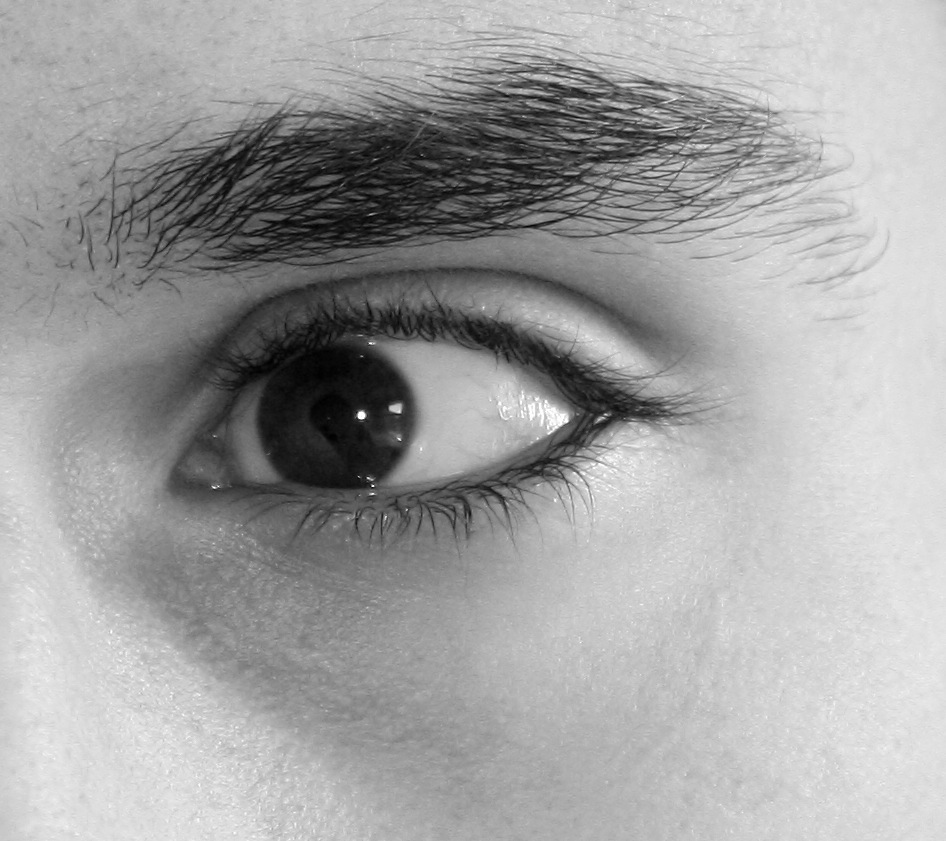}\\
\mbox{\scriptsize{(a) Grayscale periocular images}} 
\end{array}\]
\[\begin{array}{cc}
\includegraphics[width=.7in,height=.6in]{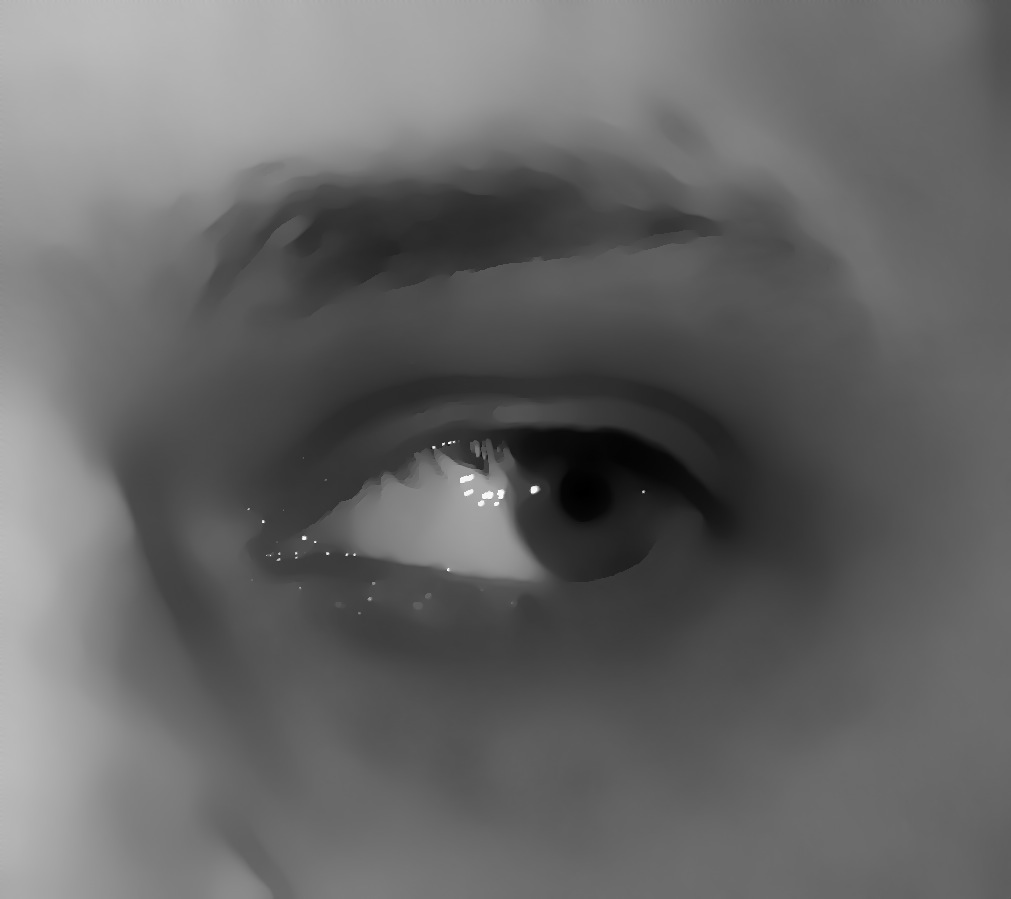}  
\includegraphics[width=.7in,height=.6in]{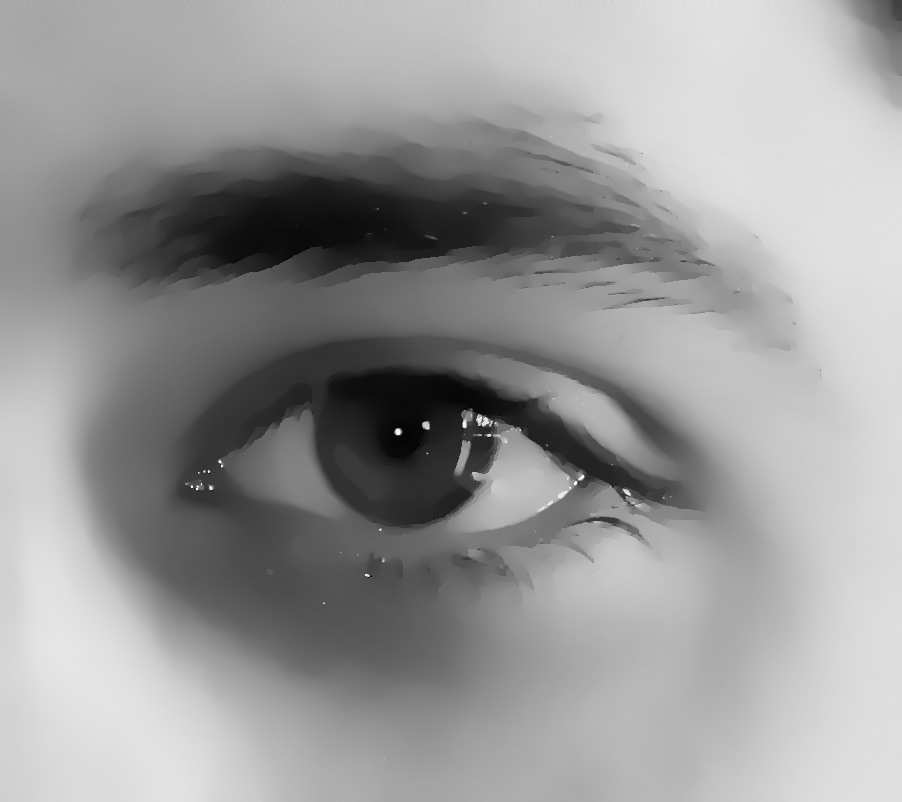}  
\includegraphics[width=.7in,height=.6in]{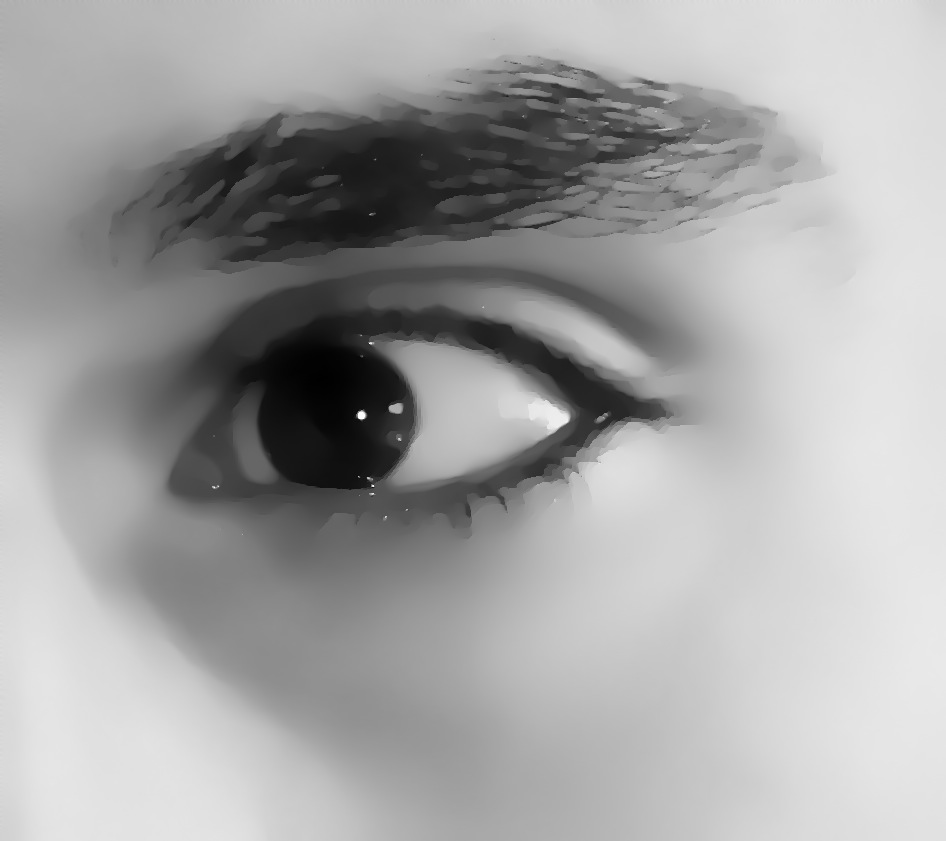}&
\includegraphics[width=.7in,height=.6in]{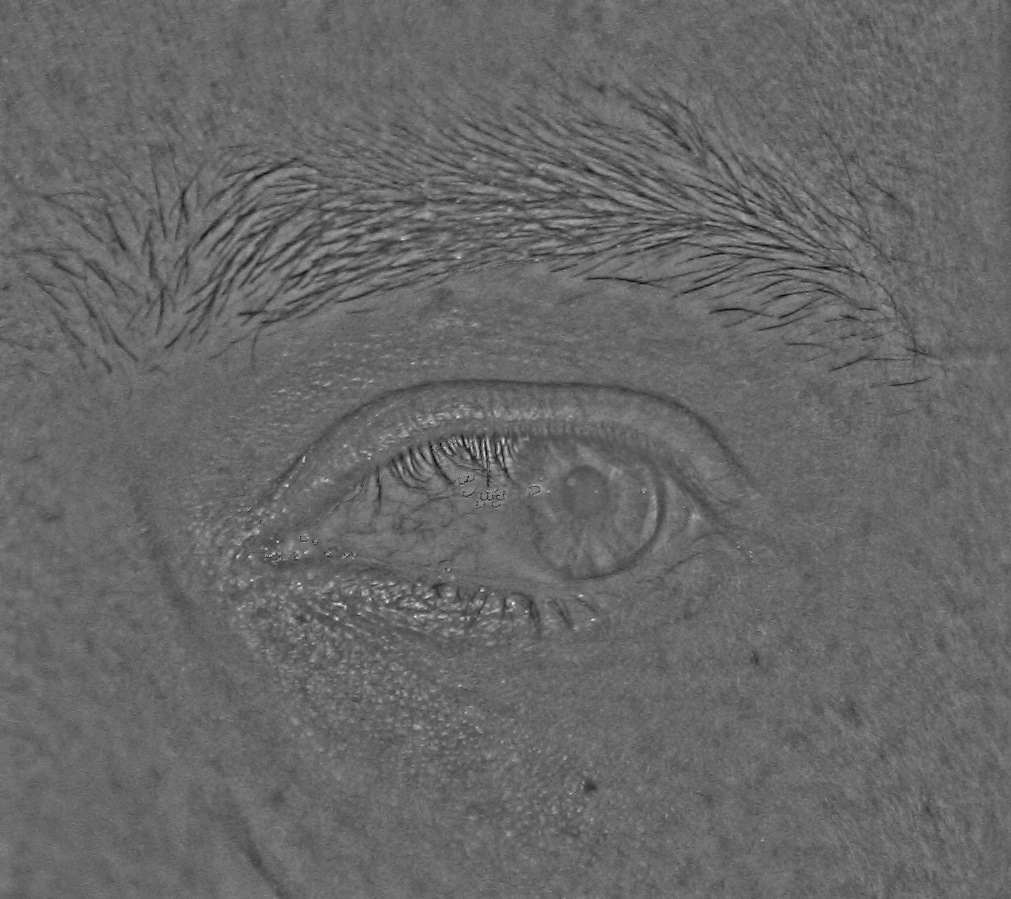}  
\includegraphics[width=.7in,height=.6in]{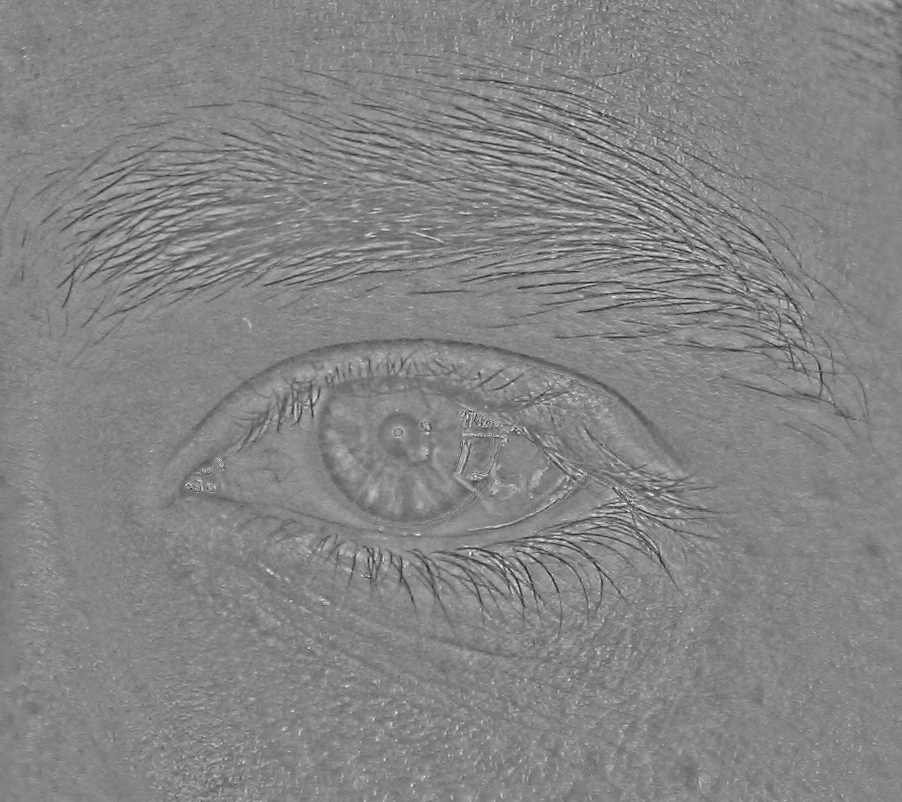}  
\includegraphics[width=.7in,height=.6in]{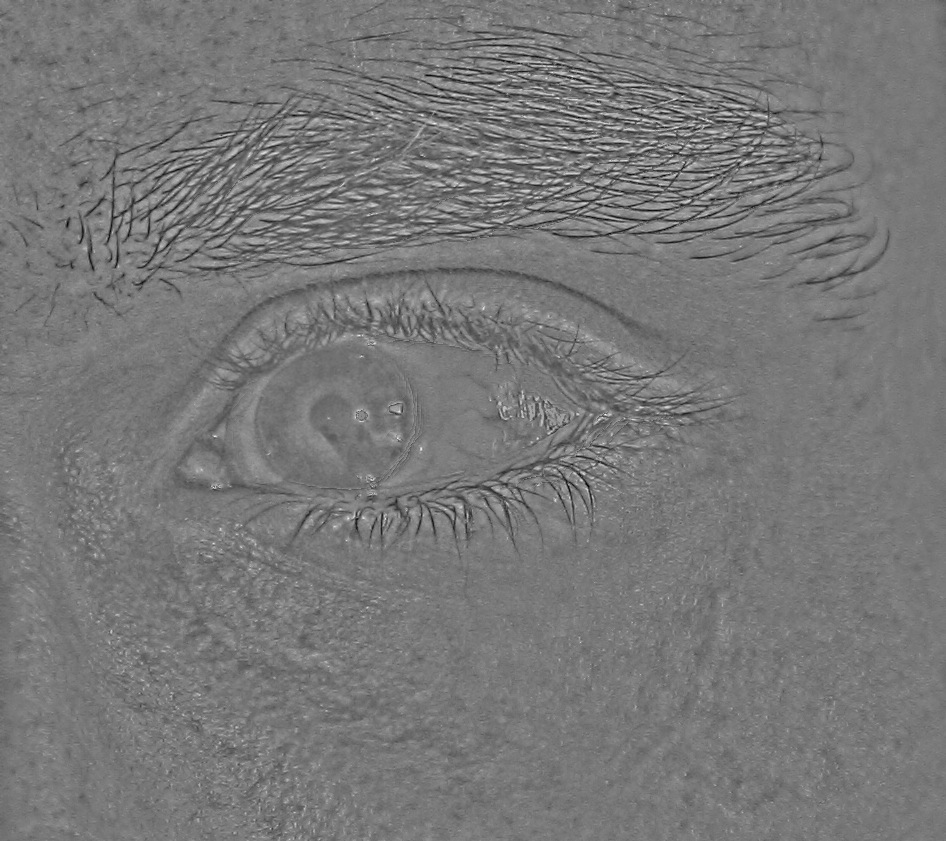}\\
\mbox{\scriptsize{(b) Cartoon Component with 80 iterations}} &
\mbox{\scriptsize{(c) Texture Component with 80 iterations}}\\
&\\
\includegraphics[width=.7in,height=.6in]{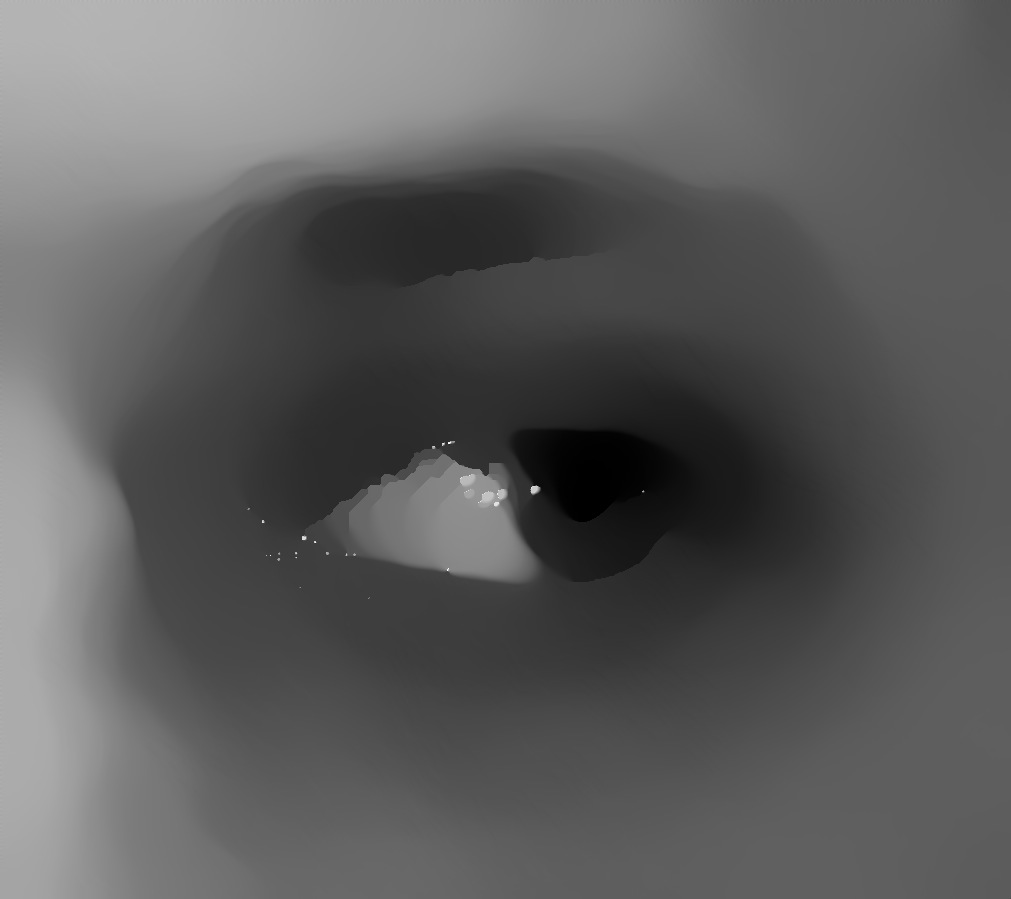}  
\includegraphics[width=.7in,height=.6in]{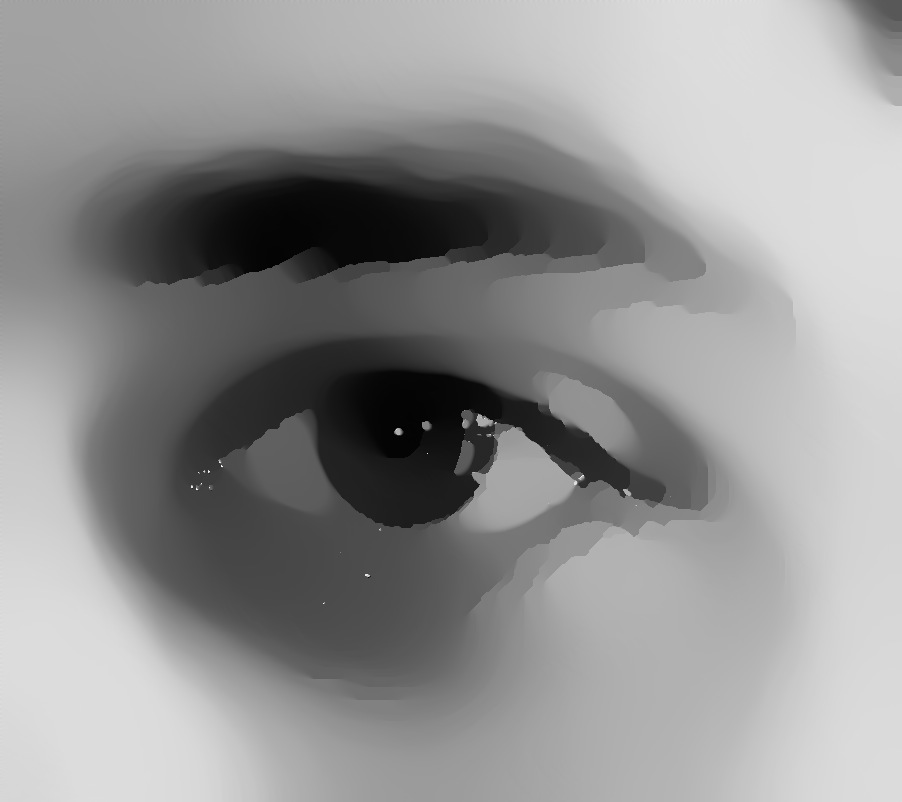}  
\includegraphics[width=.7in,height=.6in]{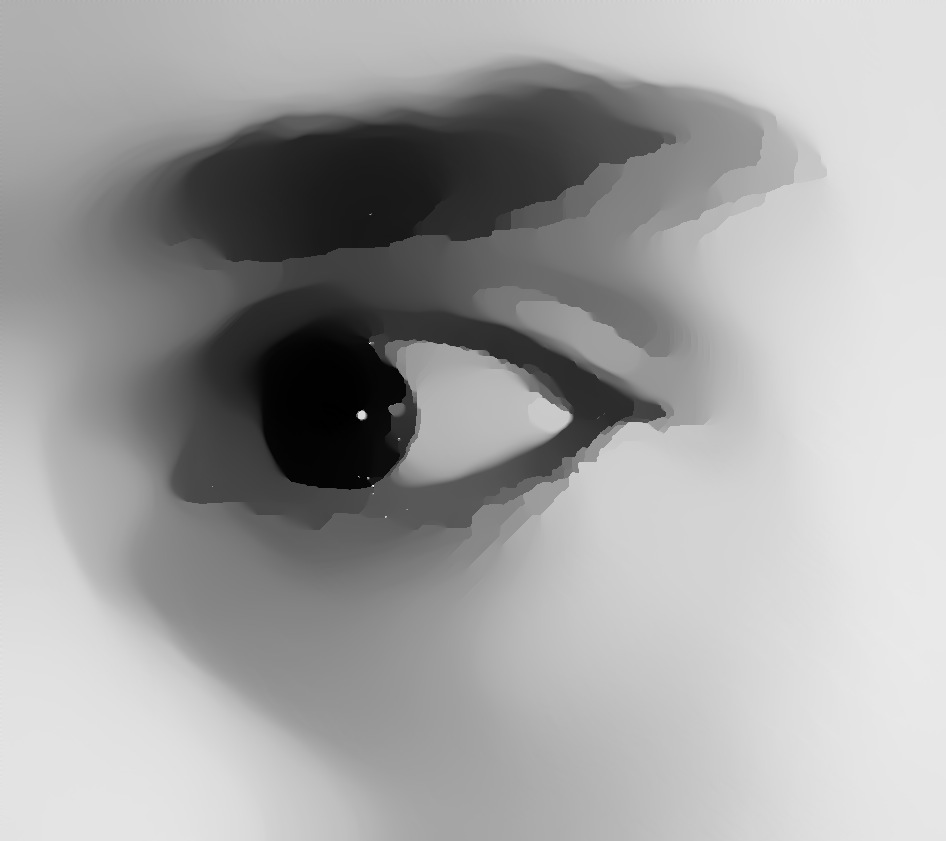}&
\includegraphics[width=.7in,height=.6in]{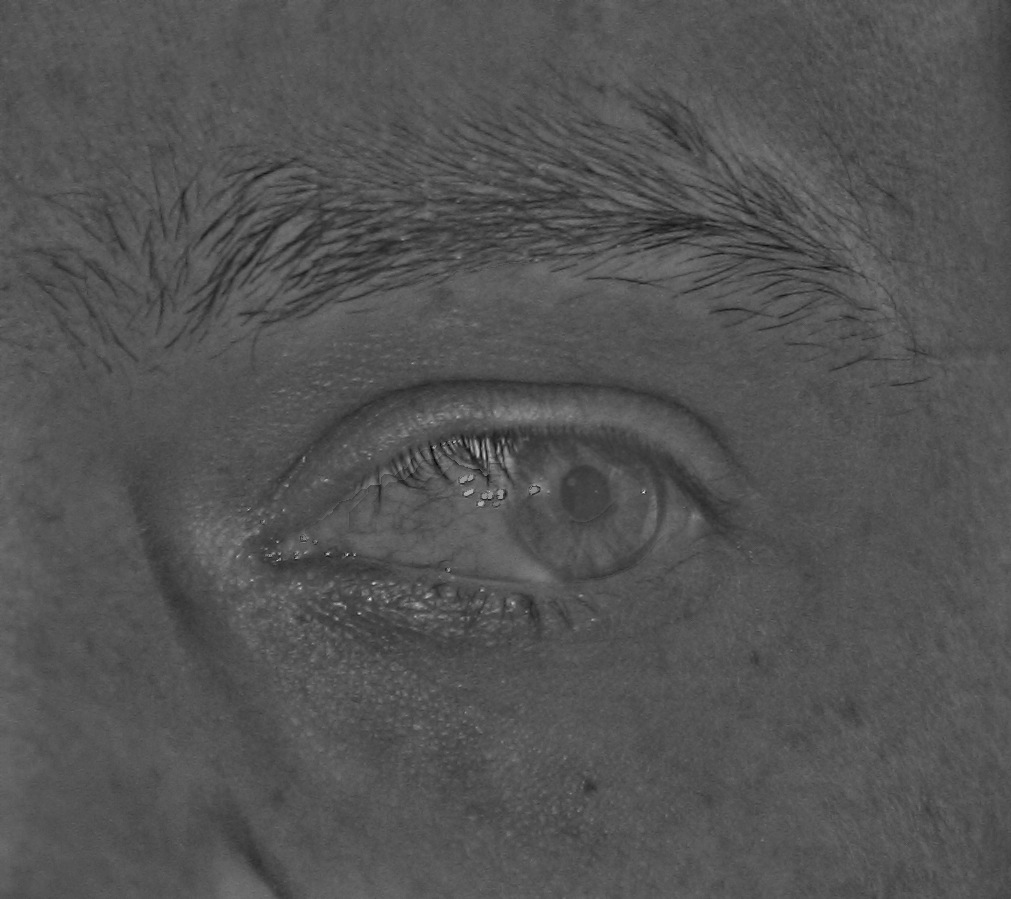}  
\includegraphics[width=.7in,height=.6in]{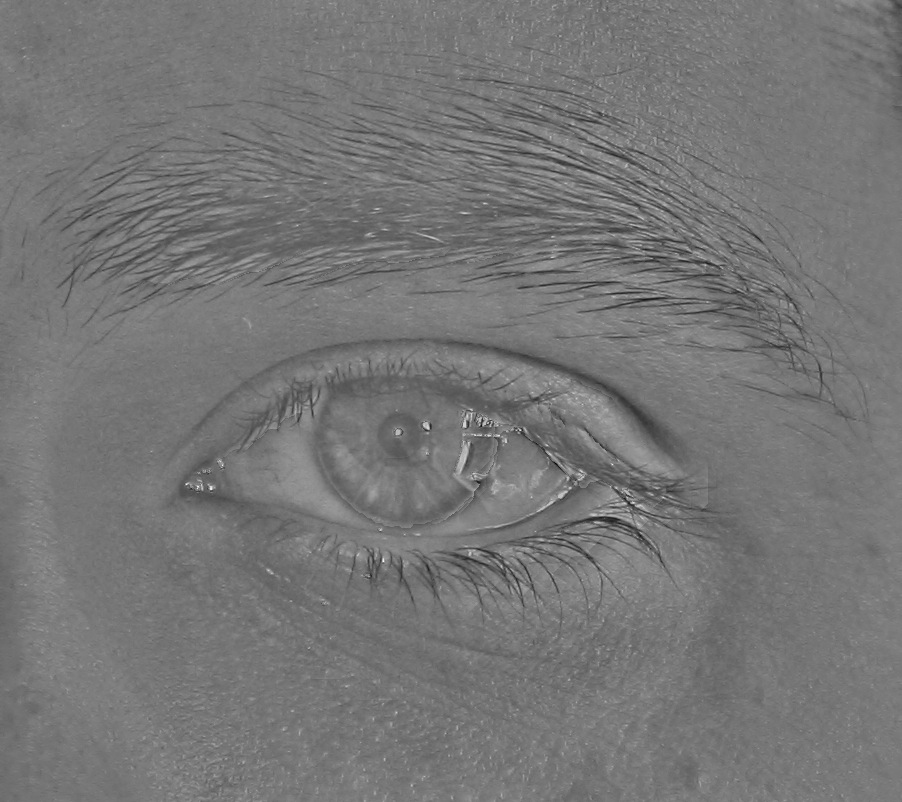}  
\includegraphics[width=.7in,height=.6in]{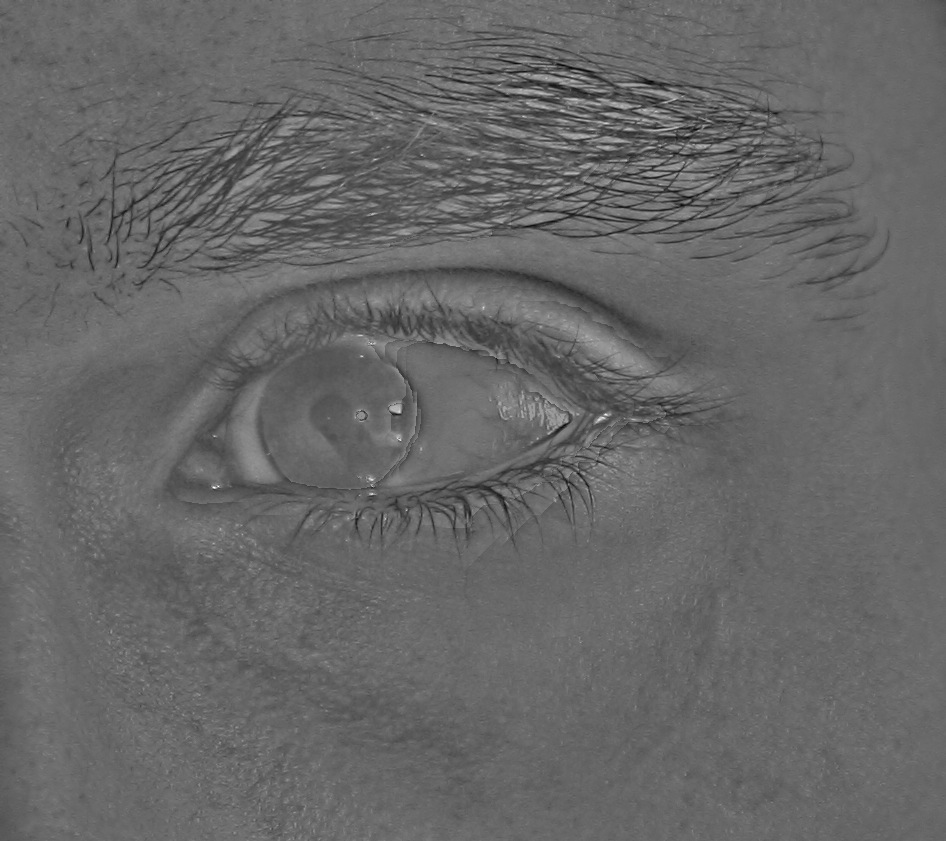}\\
\mbox{\scriptsize{(d) Cartoon Component with 400 iterations}} &
\mbox{\scriptsize{(e) Texture Component with 400 iterations}}\\
\end{array}\]
\caption{\scriptsize{Cartoon - Texture component for grayscale periocular images using a weighted TV model~\eqref{E:wTV}. 
	(a) Grayscale periocular images. 
	(b)-(c) Cartoon - Texture decomposition with 80 iterations. 
	(d)-(e) Cartoon - Texture decomposition with 400 iterations.}} 	
\label{fig:Periocular_TVdecomp}
\end{figure}
\section{Geometric and Color Spaces for Image Decomposition}\label{Proposed}
\subsection{Cartoon + Texture (CT) Space}\label{cart}

The periocular images contain cartoon (smooth) and texture parts (small scale oscillations) which can be obtained using the total variation (TV)~\cite{RO92} model effectively.  In this setting, the grayscale version of a periocular image is divided into two components representing the geometrical and texture parts. The TV based decomposition model is defined as an energy minimization problem,
\begin{equation*}\label{E:wTV}
\min\limits_{u}\left\{E^{L^{1}}_{TV}(u)=\int_{\Omega}g(\mathbf{x})|\nabla
u|\,\mathrm{d}\mathbf{x} +
\lambda\int_{\Omega}|u-I|\,\mathrm{d}\mathbf{x}\right\}
\end{equation*}
where $I$ is the input grayscale image, and $g(\mathbf{x})=\frac{1}{1+K |\nabla I |^{2}}$ is an edge indicator type function. Following~\cite{BE07} we use a splitting with an auxiliary
variable $v$ to obtain the following relaxed minimization,
\begin{eqnarray}\label{E:wTVsplit}
\min\limits_{u,v}\left\{\tilde{E}^{L^1}_{TV}(u,v)=\int_{\Omega}g(\mathbf{x})|\nabla
u|\,\mathrm{d}\mathbf{x} +
\frac{1}{2\theta}\int_{\Omega}(u+v-I)^2\,\mathrm{d}\mathbf{x}+
\lambda\int_{\Omega}|v|\,\mathrm{d}\mathbf{x}\right\}.
\end{eqnarray}
After a solution $u$ is computed, it is expected to get the
representation $I\approx u+v$, where the function $u$ represents the
geometric cartoon part, the function $v$ contains texture
information, and the function $g$ represent edges. The minimization~\eqref{E:wTVsplit} is achieved by solving the following alternating sub-problems based on the dual minimization technique:
\begin{figure}
\centering
\[\begin{array}{cc}
\includegraphics[width=.7in,height=.6in]{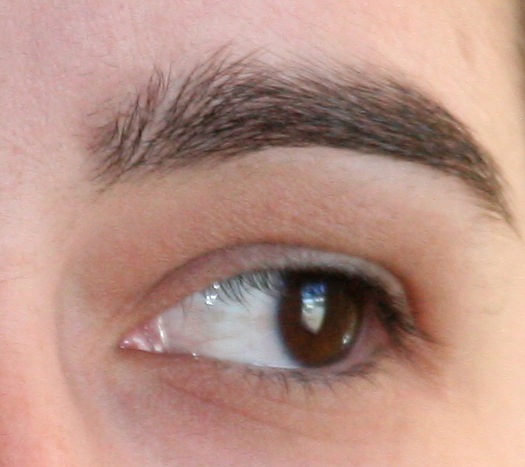} &
\includegraphics[width=.7in,height=.6in]{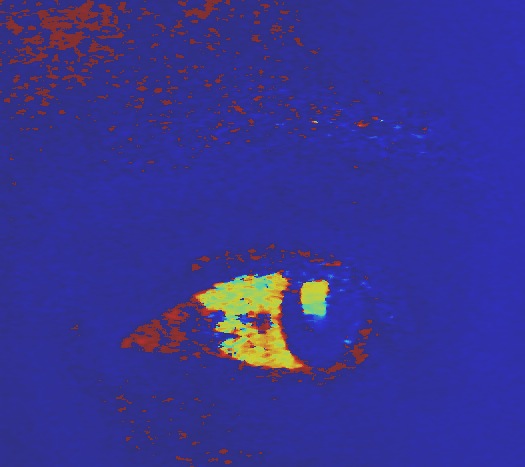}  
\includegraphics[width=.7in,height=.6in]{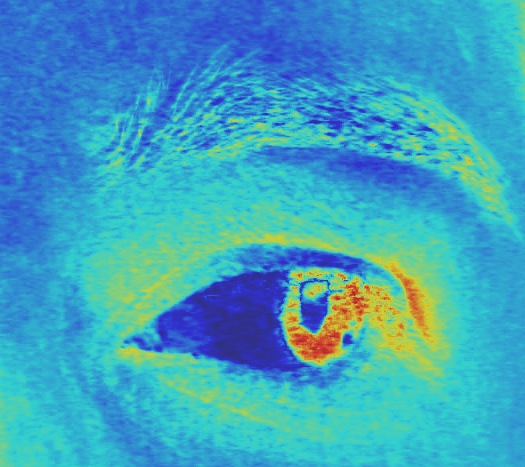}  
\includegraphics[width=.7in,height=.6in]{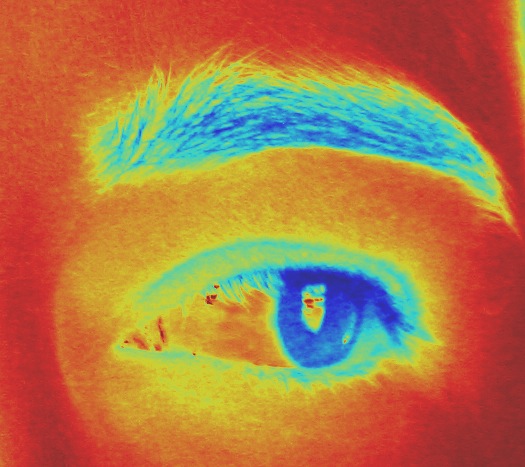}\\
\mbox{\scriptsize{(a) RGB periocular Image}} &
\mbox{\scriptsize{(b) HSV color decomposition}} \\
\end{array}\]
\[\begin{array}{c}
\includegraphics[width=.7in,height=.6in]{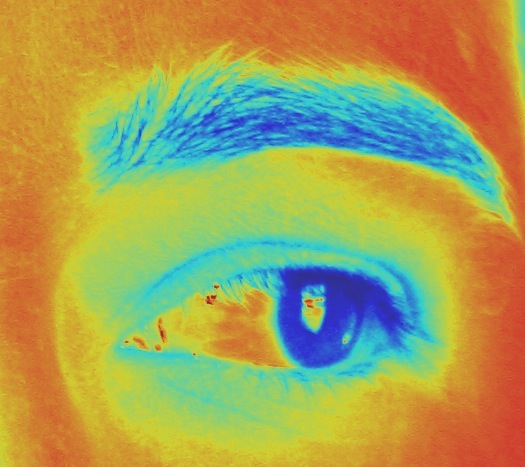}  
\includegraphics[width=.7in,height=.6in]{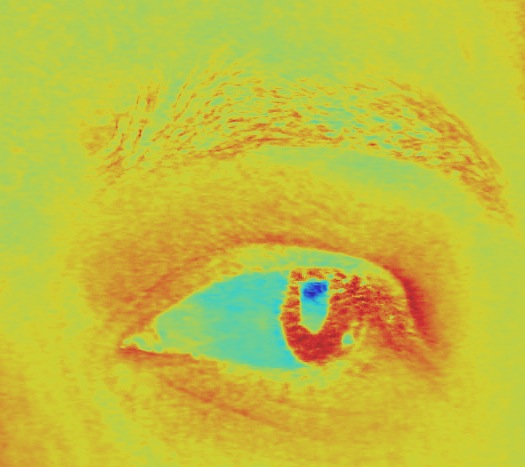}  
\includegraphics[width=.7in,height=.6in]{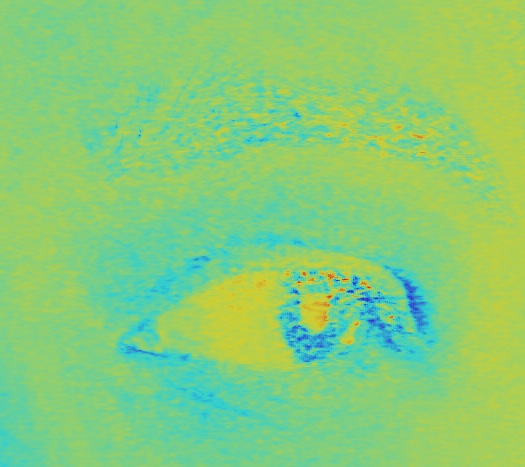}
\includegraphics[width=.7in,height=.6in]{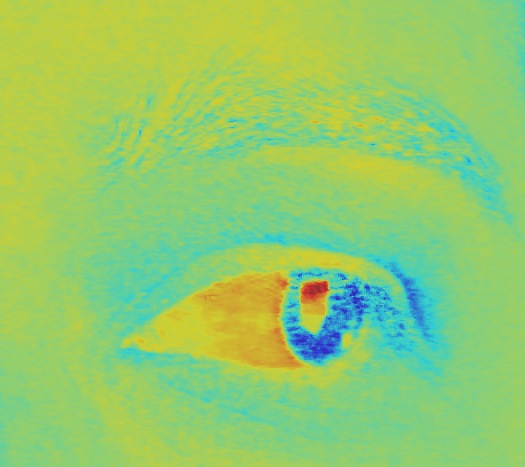}\\
\mbox{\scriptsize{(c) CB color decomposition }} \\
\end{array}\]
\caption{\scriptsize{Different color decomposition for a given periocular image. 
(a) RGB color periocular image. 
(b) HSV color decomposition. 
(c) CB color decomposition.}}	
\label{fig:Periocular_color_decomp}
\end{figure}

\begin{enumerate}
\item Fixing $v$, the minimization problem in $u$ is:
\begin{equation}\label{E:uieqn}
\min_{u}\left\{\int_{\Omega}\,g(\mathbf{x})|\nabla u |\, \mathrm{d}\mathbf{x}+
\frac{1}{2\theta}\|u+v-I\|^{2}_{L^{2}(\Omega)}\right\}.
\end{equation}
The solution of \eqref{E:uieqn} is given by $u =v - \theta\mathrm{div}\, \mathbf{p}$ where $\mathbf{p}=(p_{1},p_{2})$ satisfies $g(\mathbf{x}) \nabla(\theta\, \mathrm{div}\, \mathbf{p} -(I- v))-
|\nabla(\theta\mathrm{div}\, \mathbf{p} - (I- v)) |\mathbf{p}=0$, solved using a fixed point method: $\mathbf{p}^{0} = 0$ and iteratively 
\begin{equation*}\label{E:psoln}
\mathbf{p}^{n+1} = \frac{\mathbf{p}^{n}+\delta t \nabla(\mathrm{div}(\mathbf{p}^{n}) -
(I-v)/\theta) }{1+\frac{\delta t}{g(\mathbf{x})} |\nabla(\mathrm{div}(\mathbf{p}^{n}) - (I-v)/\theta)|}.
\end{equation*}

\item Fixing $u$, the minimization problem in $v$ is:
\begin{equation*}\label{E:vieqn}
\min_{v}\left\{\frac{1}{2\theta}\|u+v-I\|^{2}_{L^{2}(\Omega)}+\lambda\|v\|_{L^{1}(\Omega)}  \right\},
\end{equation*}
and the solution is found as 
\begin{equation*}\label{E:visoln}
v=  
\begin{cases} 
I-u-\theta\lambda & \text{if $I-u\geq \theta\lambda$},\\
I-u+\theta\lambda & \text{if $I-u\leq -\theta\lambda$},\\
0 & \text{if $|I-u|\leq \theta\lambda$}.
\end{cases}
\end{equation*}
\end{enumerate}
Figure~\ref{fig:Periocular_TVdecomp} illustrates cartoon - texture decomposition of three grayscale periocular images for different iterations. As the number of iterations we notice that the cartoon component becomes smoother and texture component picks up more oscillations.
\subsection{Color Spaces}\label{color}

For color periocular images we can obtain intensity and chromaticity decomposition which exploits color information. In computer vision there has been increasing interest in non flat image features that live on curved manifolds which are well suited for edge detection and enhancement in color and multichannel images~\cite{TSCaselles01}. The flatness concept is related to functions taking all possible values in an open set in a linear space. The chromaticity feature of color images is an example of non-flat features.  Given a color periocular image $\mathbf{I}:\Omega\rightarrow\mathbb{R}^{3}$, the RGB representation is defined by a vector with three components $\mathbf{I}=(I_{1},I_{2},I_{3})$. From the RGB color space, the chromaticity-brightness (CB) model arises by decomposing into the brightness component $\mathbf{B}:\Omega\rightarrow\mathbb{R}$ computed as $\mathbf{B}=|\mathbf{I}|$ and chromaticity components  $\mathbf{C}=(C_{1},C_{2},C_{3}):\Omega\rightarrow\mathbb{S}^{2}$ (where $\mathbb{S}^{2}$ is the unit sphere in $\mathbb{R}^{3}$) is computed by $C_{i}=I_{i}/\mathbf{B}$. We also make use of the  Hue-Saturation-Value (HSV) color space commonly used since it is believed to be more natural and is related to human perception~\cite{WStiles82}. Figure~\ref{fig:Periocular_color_decomp} illustrates CB decomposition, and HSV color space conversions of a given RGB periocular image. In our experiments we compare grayscale CT decomposition and CB, RGB and HSV color space based decompositions along with the proposed REN model.  
\section{Experiments and Discussion}\label{exper}

\begin{table}
\begin{center}
\scriptsize{
\begin{tabular}{ccc}
\hline
\multirow{2}{*}{Signal $\mathbf{\hat{x}}$} & Optimal  & Non-Optimal \\
 & Signal Recovery & Signal Recovery\\
\hline
$\text{SCI}(\mathbf{\hat{x}})>\beta\rightarrow \text{Positive}$       & \mbox{True Positive (TP)}&\mbox{False Positive (FP)} \\
$\text{SCI}(\mathbf{\hat{x}})\leq \beta\rightarrow\text{Negative}$ &\mbox{False Negative (FN)}&\mbox{True Negative (TN) }\\
\hline
\end{tabular}
}
\end{center}
\caption{\scriptsize{Types of errors, according to the SCI value and the sparse signal reconstruction following Wright {\it et. al.}~\cite{WYGSM09} and Pillai~\etal~\cite{PPCR11} models.}}
\label{tab:Classifier}
\end{table}
\subsection{Comparison Terms: Related Periocular Recognition Methods}\label{comparison}

The use of the periocular region is found to be useful on unconstrained scenarios~\cite{santos2013}. The exploration of the periocular region as a biometric trait started with Park~\etal's pioneering approach~\cite{park2009}, who performed local and global feature extraction. Images were aligned to take advantage of iris location, in order to define a $7\times 5$ \ac{ROI} grid.  Patches were encoded by applying two well known distribution-based descriptors, \ac{LBP}~\cite{ojala1994} and \ac{HOG}~\cite{dalal2005}, quantized into 8-bin histograms. Finally, they merged all histograms into a single-dimension array containing both texture and shape information, and matching was carried out based on the Euclidean distance. For the local analysis, authors employed \ac{SIFT}~\cite{lowe2004}. The reported performance was fairly good, showing periocular fitness for recognition purposes, and further analysis was held on noise factors impact on performance~\cite{park2011}. 

Recently, various extensions and improvements based on Park et al work~\cite{park2009} has been carried out. Miller~\etal~\cite{miller2010b} presented an analysis which focused on periocular skin texture, taking advantage of \ac{ULBP}~\cite{ojala2002} to achieve improved rotation invariance with uniform patterns and finer quantization of the angular space. Their work was extended by Adams~\etal~\cite{adams2010}, who proposed using \ac{GEC} to optimize feature set.  Juefei-Xu~\etal~\cite{juefei-xu2010} used multiple local and global feature extraction techniques such as Walsh transforms and Laws' masks, \ac{DCT}, \ac{DWT}, Force Fields, \ac{SURF}, Gabor filters and \ac{LoG}. In their later work~\cite{juefei-xu2011} efforts were made to compensate aging degradation effects on periocular performance. The possibility of score level fusion with other biometric traits was also addressed, for example in iris recognition~\cite{woodard2010}. Bharadwaj~\etal~\cite{bharadwaj2010} proposed the fusion of \ac{ULBP} with five perceptual dimensions, usually applied as scene descriptors: naturalness, openness, roughness, expansion and ruggedness -- GIST~\cite{oliva2001}. In their approach the images were pre-processed with with Fourier transform for local contrast normalization, and then a spacial envelope computed with a set of Gabor filters (4 scales $\times$ 8 orientations).
On the final stage, $\chi^2$ distance was used to match the feature arrays, and results fused with a weighted sum.

\begin{table*}
\centering
\scriptsize{
\begin{tabular}{cccccccc}
\hline
Mehtod&Feature& sens. & far & acc. & thres. & AUC & EER \\
\hline
\multirow{6}{*}{REN (Proposed)} &Grayscale (SRC) & 90.05 & 8.55  & 90.99 & 0.1553 &0.9643 &0.0904\\
                                                 &Texture (SRC)                 & 92.10 & 1.89  & 92.40 & 0.0756 &0.9756                       &0.0589\\
                                                 &CT (Fusion)                      & 99.90 & 7.18  & 98.77 & 0.1641 &\underline{0.9994}  &\underline{0.0018}\\
                                                
                                                 &CB (Fusion)	                    & 99.82 & 7.37  & 98.49 & 0.2333 &0.9992  &0.0061\\
                                                 &RGB (Fusion)		 & 99.83 & 4.11  & 99.31 & 0.1670  &0.9990                       &0.0020\\
                                                 &HSV (Fusion)                   & 99.83 & 2.13  & 99.57 & 0.1832  &0.9991  &0.0019\\
\hline
\multirow{1}{*}{Wright2009} &Grayscale (SRC)                 & 84.70 & 9.59  & 85.14 & 0.05642 &0.9307                                   &0.1529\\                                                 
\hline
\multirow{4}{*}{Park2009}            & LBP  & 80.70& 9.99 & 86.90 & 0.7468 & 0.9189 & 0.1553 \\
                                                          & HOG & 69.29& 9.99 & 83.11 & 0.6421 & 0.8656 & 0.2088 \\
                                                          & SIFT  & 86.00& 9.36 & 88.96 & 0.0477 & 0.9453 & 0.1232 \\
                                                       & Fusion & 90.58& 9.99 & 90.21 & 0.1052 & 0.9564 & 0.0954\\
\hline
\multirow{3}{*}{Bharadwaj2010} & GIST  & 75.56& 9.99 & 85.20 & 0.7623 & 0.8927 & 0.1846 \\
                                                          & ULBP & 85.82& 9.99 & 88.61 & 0.8673 & 0.9259 & 0.1311 \\
                                                         & Fusion & 83.96& 9.99 & 88.00 & 0.8008 & 0.9235 & 0.1386 \\
\hline

\end{tabular}
}
\caption{\scriptsize{AUC and EER values, as well as the best sensitivity for $\text{far} \leq 10\%$ for left side periocular images. The Underline fonts indicate the best model observed.}}
\label{tab:ROC_SV1}
\end{table*}
\subsection{Performance Measures}

Images were down-sampled to $10\times 9$ pixels and stored in ``png'' format. The resulting sensitivity and specificity values were considered, obtaining the Receiver Operating Characteristic curves (ROC). In this case, given a signal $\mathbf{\hat{x}}$, if $\text{SCI}(\mathbf{\hat{x}})>\beta$, the classifier outputs a positive response (P), otherwise a negative (N) result. For a fixed $\beta$, the sensitivity corresponds to the proportion of signals correctly detected by the SRC algorithm, whereas specificity counts the proportion for which the corresponding SCI values are bellow $\beta$, where $\beta$ is an accepted threshold value. 
\begin{equation*}\label{Expe:eq2}
\text{sensitivity} = \frac{\mbox{\#TP}}{\mbox{\#TP + \mbox{\# FN}}}\quad \mbox{and}\quad \text{specificity} = \frac{\mbox{\# TN}}{\mbox{\# TN + \mbox{\#FP}}},
\end{equation*} 
where TP, FP, TN and FN correspond to the True Positive, False Positive, True Negative and False Negative, respectively. Table~\ref{tab:Classifier} summarizes these notions, combining the different classes of periocular signals and their relation with the classifier induced by the minimal reconstruction error and the accumulated SCI value. 
The overall accuracy is given by:
\begin{equation*}
\text{accuracy}=\frac{\mbox{\# TN}+\mbox{\# TP }}{\mbox{\# TN}+\mbox{\# FP } + \mbox{\# TP}+\mbox{\# FN}}.
\end{equation*}
In a ROC plot, the optimal recognition method would yield a point in the upper-left corner, corresponding to full sensitivity (no false negatives) and full specificity (no false positives).  The statistical correlation between the outputs given by each channels considered in our method was also assessed. Considering that eventual dependences will be linear, the Pearson's sample correlation was used for that purpose. Given a pair of samples, the correlation coefficient is given by:
\begin{equation*}
r(\mathbf{\hat{x}}^{(1)},\mathbf{\hat{x}}^{(2)})=\displaystyle\frac{1}{n-1}\sum_{i=1}^{n}\left(\frac{\hat{x}^{(1)}_{i}-\bar{\mathbf{x}}^{(1)}}{\sigma_{\mathbf{\hat{x}}^{(1)}}}\right)\left(\frac{\hat{x}^{(2)}_{i}-\bar{\mathbf{x}}^{(2)}}{\sigma_{\mathbf{\hat{x}}^{(2)}}}\right),
\end{equation*}
where $\hat{x}^{(1)}_{i}$, $\hat{x}^{(2)}_{i}$ denote the systems outputs, $\bar{\mathbf{x}}^{(1)}$, $\bar{\mathbf{x}}^{(2)}$ are the sample means and $\sigma_{\mathbf{\hat{x}}^{(1)}}$, $\sigma_{\mathbf{\hat{x}}^{(2)}}$ the standard deviations. 

\subsection{Results}\label{left}

For our first experiment, we focus on only left side periocular images. Six samples from $150$ different subjects were used, such that one image per class was randomly chosen as probe and the remaining five samples  included in the dictionary.  Experiments were repeated, changing the image used as probe (per subject). Hence, $100$ dictionaries with dimension $90\times 750$ were considered, each one tested in $150$ probe samples. 

Results are summarized in Table~\ref{tab:ROC_SV1} in terms of true and false positive rates where the best sensitivity (sens.) and corresponding accuracy (acc.) for far (=1-specificity) $\leq 10\%$ have been computed for various schemes and models studied here. The proposed reweighed elastic net demonstrates to be superior than the original SRC approach over grayscale impulses. In this case the area under the curve (AUC) and the equal error rate (EER) are equal to 0.9643 and 0.0904 for our model, against 0.9307 and 0.1529 produced by the original SRC model. The proposed models approximates more to the \emph{optimal performance} point (complement of $\text{specificity}=0$, $\text{sensitivity}=1$). For the REN approach applied to the grayscale and the texture components alone the minimal distance from the ROC values to the $(0,1)$ point was of $0.1511$ and $0.0812$ respectively, while the value $0.1805$ was observed for the classical SRC model. In relation to other image representation components, the minimal distance from the ROC values to the $(0,1)$ point was of $0.0022$ $0.0106$, $0.0023$, $0.0019$ for the CT, CB, RGB and HSV spaces. 

\begin{figure*}
\centering
\subfigure[\scriptsize{ROC curves - Left Side}]{\includegraphics[width=3.9cm,height=3.9cm]{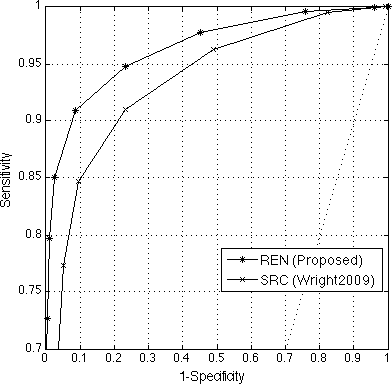}}
\subfigure[\scriptsize{ROC curves - Left Side} ]{\includegraphics[width=3.9cm,height=3.9cm]{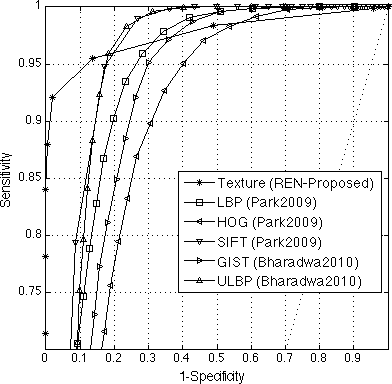}}
\subfigure[\scriptsize{ROC curves - Left Side} ]{\includegraphics[width=3.9cm,height=3.9cm]{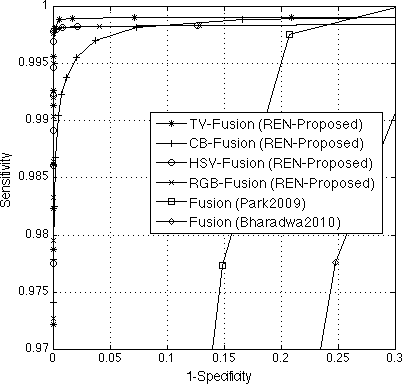}}\\
\caption{\scriptsize{ROC curves for periocular images recognition. (a) ROC curves for the original REN approach and the SRC model and the REN approach. (b) ROC curves for the REN approach applied to the texture components together with different features extrated by Park~\etal~\cite{park2009} and Bharadwaj~\etal~\cite{bharadwaj2010}. (c) ROC curves for the REN model applied to the proposed fusion over the different geometry and color spaces, as well as the fusion implemented in Park~\etal~\cite{park2009} and Bharadwaj~\etal~\cite{bharadwaj2010}. }}
\label{fig:ROC_1}
\end{figure*}

Comparisons have been carried out by implementing the well known models of Park~\etal~\cite{park2009} and Bharadwaj~\etal~\cite{bharadwaj2010}. Even both models make full use of local and global periocular information to perform recognition, they have shown not to improve better than our approach. In our experiments, we have compared the AUC as well as the EER values in the case REN model approach uses the texture periocular components as feature extraction, against those features used in Park~\etal~\cite{park2009} and Bharadwaj~\etal~\cite{bharadwaj2010}. For the comparison models the highest AUC is equal to 0.9564 and lowest EER is equal to 0.0954 when applying their fusion techniques. Meanwhile, using the texture information provided by the cartoon - texture space, our model got the values 0.9756 and 0.0589 for the AUC and EER, respectively. Our fusion method using different spaces completely describing the geometry and color periocular feature have also shown to reached great  statistical values in comparison to those values got it by Park~\etal~\cite{park2009} and Bharadwaj~\etal~\cite{bharadwaj2010} approaches. In this case, the highest AUC and the lowest EER values are given by the CT space with values 0.9994  and 0.0018, see Figure \ref{fig:ROC_1}.

\begin{table*}
\centering
\scriptsize{
\begin{tabular}{ccccccc}
\hline
& Grayscale & Texture &CT & CB & RGB & HSV   \\
\hline
Grayscale                                 & 1 & 0.7173  & 0.5331 & 0.4272 & 0.7139  & 0.6230   \\
Texture                                      & -  & 1            & 0.6041 & 0.6285 & 0.9776  & 0.9043   \\
CT                                               & -  & -             & 1           & 0.2206 & 0.6134  & 0.6146   \\
CB	                                           & -  & -             & -            & 1           & 0.6146  & 0.7213   \\
RGB		                                  & -  & -             & -            & -           & 1             & 0.9180   \\
HSV		                                  & -  & -             & -            & -           & -               & 1 \\
\\\hline
\end{tabular}
}
\caption{\scriptsize{Pearson's sample correlation coefficients between the left side responses given by the recognition algorithms using the REN model with various components studied here.}}
\label{tab:Perason's sample1}
\end{table*}

As it can be observed from Table~\ref{tab:Perason's sample1}, the proposed REN model applied to grayscale in texture setting are in high correlation when compared to the signals recovered in the CT, RGB and HSV spaces. The result is due to the high accuracy rates achieved over these image representations. Similarly the signals recovered in the CB space are in low correlation with the signals lying in the grayscale setting and the CT space, and in high correlation with the signals computed over texture domain alone. This is because the chromaticity components lying in the unit sphere $\mathbb{S}^{2}$ have the advantage of depicting nonlinear features in different directions and therefore both strong and weak edges are distributed and represented along chromaticity components. Also, it should be noted the strong correlation between the outputs given by the fusion model when using exclusively color components. This is also can be explained, as the skin region comprises a large majority of the periocular region (see Figure~\ref{fig:Periocular_color_decomp}). It is particularly interesting to observe that the positive (and small) correlation values between the signals are obtained when using different color spaces representation, pointing for a complementarity that might contribute for the outperforming results of the method proposed in this paper. Although the CT space produces good recognition rates, its computed signals are in low correlation with respect to other signals over different domains, owing to the fact that CT space is given by geometric information in case of cartoon component, whereas weak and strong edges describe texture components, see Figure~\ref{fig:Periocular_TVdecomp}.         

\section{Conclusions}\label{conc}

This paper described a novel re-weighted elastic net (REN) model that improves the sparsity of representations, which is particularly useful for classification purposes. The weights are used for penalizing different coefficients in the REN-penalty. In addition, theoretical existence results have been proved for the REN minimization problem, mainly emphasizing our approach is good in the sense it performs as well as if the true underlying model were given in advance. As far as numerical approximation is concerned, the REN model is expressed as a quadratic programing (QP) expediting the implementation of the proposed gradient projection (GP) algorithm and providing good results.  

To empirically validate the proposed method, we used the \emph{periocular region} which is an emerging biometric trait with high potential to handle data acquired under uncontrolled conditions.  From this perspective, the main novelty of the proposed scheme is to fuse multiple sparse representations, associated with various spaces from different domains in geometry and color. This allowed us to faithfully handle distortions in periocular images such as blur and occlusions. Our experiments were carried out in the highly challenging images of the UBIRIS.v2 dataset, and allowed us to observe consistent improvements in performance, when compared to the classical sparse representation model, and state-of-the-art periocular recognition algorithms.

\bibliographystyle{plain}
\bibliography{BTAS_reference,periocular} 

\begin{thebibliography}{10}

\bibitem{adams2010}
J.~Adams, D.L. Woodard, G.~Dozier, P.~Miller, K.~Bryant, and G.~Glenn.
\newblock Genetic-based type ii feature extraction for periocular biometric
  recognition: Less is more.
\newblock In {\em Pattern Recognition (ICPR), 2010 20th International
  Conference on}, pages 205 --208, August 2010.

\bibitem{bharadwaj2010}
S.~Bharadwaj, H.S. Bhatt, M.~Vatsa, and R.~Singh.
\newblock Periocular biometrics: When iris recognition fails.
\newblock In {\em Biometrics: Theory Applications and Systems (BTAS), 2010
  Fourth IEEE International Conference on}, pages 1 --6, September 2010.

\bibitem{BE07}
X.~Bresson, S.~Esedoglu, P.~Vandergheynst, J.~Thiran, and S.~Osher.
\newblock Fast global minimization of the active contour/snake model.
\newblock {\em Journal of Mathematical Imaging and Vision}, 28(2):151--167,
  2007.

\bibitem{CRT06}
E.~Cand\`{e}s, J.~Romberg, and T.~Tao.
\newblock Stable signal recovery from incomplete and inaccurate measurements.
\newblock {\em Communications on Pure and Applied Mathematics},
  59(8):1207--1223, 2006.

\bibitem{CTao05}
E.~Candes and T.~Tao.
\newblock Decoding by linear programming.
\newblock {\em IEEE Transactions on Information Theory}, 51(12):4203--4215,
  2005.

\bibitem{CTao07}
E.~Candes and T.~Tao.
\newblock The dantzig selector: statistical estimation when $p$ is much larger
  than $n$.
\newblock {\em The Annals of Statistics}, 35(6):2392--2404, 2007.

\bibitem{CWBoyd08}
E.~Cand\`{e}s, M.~Wakin, and Stephen~P. Boyd.
\newblock Enhancing sparsity by reweighted $\ell^{1}-$ minimization.
\newblock {\em Journal of Fourier Analysis and Applications}, 14(5):877--905,
  2008.

\bibitem{CDS98}
S.~Chen, D.~Donoho, and M.~Saunders.
\newblock Atomic decomposition by basis pursuit.
\newblock {\em SIAM Journal on Scientific Computing}, 20(1):33--61, 1998.

\bibitem{dalal2005}
N.~Dalal and B.~Triggs.
\newblock Histograms of oriented gradients for human detection.
\newblock In {\em In CVPR}, pages 886--893, 2005.

\bibitem{Denoho706}
D.~Donoho.
\newblock For most large underdetermined systems of equations, the minimal
  $\ell^{1}$-norm near-solution approximates the sparsest near-solution.
\newblock {\em Communications on Pure and Applied Mathematics}, 59(7):907--934,
  2006.

\bibitem{FLv08}
J.~Fan and J.~Lv.
\newblock Sure indepence screening for ultrahigh dimensional feature space.
\newblock {\em Journal of the Royal Statistical Society: Series B (Statistical
  Methodology)}, 70(5):849--911, 2008.

\bibitem{FNW07}
M.~Figueiredo, R.~Nowak, and S.~Wright.
\newblock Gradient projection for sparse reconstruction: Application to
  compressed sensing and other inverse problem.
\newblock {\em IEEE Journal of Selected Topics in Signal Processing},
  1(4):586--597, 2007.

\bibitem{Fuchs}
J.~J. Fuchs.
\newblock Multipath time-delay detection and stimation.
\newblock {\em IEEE Transactions on Signal Processing}, 47(1):237--243, 1999.

\bibitem{HZhang10}
D.~Hong and F.~Zhang.
\newblock Weigthed elastic net model for mass spectrometry image processing.
\newblock {\em Mathematical Modelling of Natural Phenomena}, 5(3):115--133,
  2010.

\bibitem{JFR07}
A.~K. Jain, P.~Flynn, and A.~Ross (Eds).
\newblock {\em Handbook of biometrics}.
\newblock Springer-Verlag, New York, USA, 2007.

\bibitem{JYu10}
J.~Jia and B.~Yu.
\newblock On model salection consistency of the elastic net when $p\gg n$.
\newblock {\em Statistica Sinica}, 20:595--611, 2010.

\bibitem{JCL10}
R.~Jiang, D.~Crookes, and N.~Lie.
\newblock Face recognition in global harmonic subspace.
\newblock {\em IEEE Transactions on Information Forensics and Security},
  5(3):416--424, 2010.

\bibitem{juefei-xu2010}
F.~Juefei-Xu, M.~Cha, J.L. Heyman, S.~Venugopalan, R.~Abiantun, and
  M.~Savvides.
\newblock Robust local binary pattern feature sets for periocular biometric
  identification.
\newblock In {\em Biometrics: Theory Applications and Systems (BTAS), 2010
  Fourth IEEE International Conference on}, pages 1 --8, sept. 2010.

\bibitem{juefei-xu2011}
F.~Juefei-Xu, K.~Luu, M.~Savvides, T.D. Bui, and C.Y. Suen.
\newblock Investigating age invariant face recognition based on periocular
  biometrics.
\newblock In {\em Biometrics (IJCB), 2011 International Joint Conference on},
  pages 1 --7, October 2011.

\bibitem{KMarch07}
S.~Kang and R.~March.
\newblock Variational models for image colorization via chromaticity and
  brightness decomposition.
\newblock {\em IEEE Transactions on Image Processing}, 16(9):2251--2261, 2007.

\bibitem{lowe2004}
D.G. Lowe.
\newblock Distinctive image features from scale-invariant keypoints.
\newblock {\em Int. J. Comput. Vision}, 60(2):91--110, November 2004.

\bibitem{miller2010b}
P.E. Miller, A.W. Rawls, S.J. Pundlik, and D.L. Woodard.
\newblock Personal identification using periocular skin texture.
\newblock In {\em Proceedings of the 2010 ACM Symposium on Applied Computing},
  SAC '10, pages 1496--1500, New York, NY, USA, 2010. ACM.

\bibitem{ojala1994}
T.~Ojala, M.~Pietikainen, and D.~Harwood.
\newblock Performance evaluation of texture measures with classification based
  on kullback discrimination of distributions.
\newblock In {\em Pattern Recognition, 1994. Vol. 1 - Conference A: Computer
  Vision amp; Image Processing., Proceedings of the 12th IAPR International
  Conference on}, volume~1, pages 582 --585 vol.1, October 1994.

\bibitem{ojala2002}
T.~Ojala, M.~Pietikainen, and T.~Maenpaa.
\newblock Multiresolution gray-scale and rotation invariant texture
  classification with local binary patterns.
\newblock {\em Pattern Analysis and Machine Intelligence, IEEE Transactions
  on}, 24(7):971 --987, July 2002.

\bibitem{oliva2001}
A.~Oliva and A.~Torralba.
\newblock Modeling the shape of the scene: A holistic representation of the
  spatial envelope.
\newblock {\em International Journal of Computer Vision}, 42:145--175, 2001.

\bibitem{PJ10}
U.~Park and A.~K. Jain.
\newblock Face matching and retrieval using soft biometrics.
\newblock {\em IEEE Transactions on Information Forensics and Security},
  5(3):406--415, 2010.

\bibitem{PJRJ11}
U.~Park, R.~R. Jillela, A.~Ross, and A.~K. Jain.
\newblock Periocular biometircs in the visible spectrum.
\newblock {\em IEEE Transactions on Information Forensics and Security},
  6(1):96--106, 2011.

\bibitem{PRJ09}
U.~Park, A.~Ross, and A.~K. Jain.
\newblock Periocular bimetrics in the visible spectrum: A feasibility study.
\newblock In {\em IEEE 3rd International Conference on Biometrics: Theory,
  Applications, and Systems}, pages 153--158, Virginia, USA, 2009.

\bibitem{park2011}
Unsang Park, R.R. Jillela, A.~Ross, and A.K. Jain.
\newblock Periocular biometrics in the visible spectrum.
\newblock {\em Information Forensics and Security, IEEE Transactions on},
  6(1):96 --106, 2011.

\bibitem{park2009}
Unsang Park, A.~Ross, and A.K. Jain.
\newblock Periocular biometrics in the visible spectrum: A feasibility study.
\newblock In {\em Biometrics: Theory, Applications, and Systems, 2009. BTAS
  '09. IEEE 3rd International Conference on}, pages 1 -- 6, September 2009.

\bibitem{PPCR11}
J.~K. Pillai, V.~M. Patel, R.~Chellappa, and N.~K. Ratha.
\newblock Secure and robust iris recognition using random projections and
  sparse representations.
\newblock {\em IEEE Transactions on Pattern Analysis and Machine Intelligence},
  33(9):1877--1893, 2011.

\bibitem{PA12IEEE}
H.~Proen\c{c}a and L.~Alexandre.
\newblock Toward covert iris biometric recognition: experimental results from
  the {NICE} contests.
\newblock {\em IEEE Transactions on Information Forensics and Security},
  7(2):798--808, 2012.

\bibitem{ubiris2}
H.~Proença, S.~Filipe, R.~Santos, J.~Oliveira, and L.A. Alexandre.
\newblock The ubiris.v2: A database of visible wavelength iris images captured
  on-the-move and at-a-distance.
\newblock {\em Pattern Analysis and Machine Intelligence, IEEE Transactions
  on}, 32(8):1529 --1535, August 2010.

\bibitem{RO92}
L.~Rudin, S.~Osher, and E.~Fatemi.
\newblock Nonlinear total variation based noise removal algorithms.
\newblock {\em Physica D}, 60(1--4):259--268, 1992.

\bibitem{santos2013}
G.~Santos and H.~Proença.
\newblock Periocular biometrics: An emerging technology for unconstrained
  scenarios.
\newblock In {\em Proceedings of the {IEEE Symposium on Computational
  Intelligence in Biometrics and Identity Management -- CIBIM 2013}}, pages
  14--21, April 2013.

\bibitem{SZZanni03}
T.~Serafini, G.~Zanghirati, and L.~Zanni.
\newblock Gradient projection methods for large quadratic programs and
  applications in training support vector machines.
\newblock {\em Optimization Methods and Software}, 20(2-3):353--378, 2003.

\bibitem{SPNChellappa13}
S.~Shekhar, V.~M. Patel, N.~M. Nasrabadi, and R.~Chellappa.
\newblock Joint sparse representation for robust multimodal biometrics
  recognition.
\newblock {\em IEEE Transactions on Pattern Analysis and Machine Intelligence},
  10(99), 2013.

\bibitem{SJ10}
R.~Sznitman and B.~Jedynak.
\newblock Active testing for face detection and localization.
\newblock {\em IEEE Transactions on Pattern Analysis and Machine Inteligence},
  32(10):1914--1920, 2010.

\bibitem{TSCaselles01}
B.~Tang, G.~Sapiro, and V.~Caselles.
\newblock Color image enhancement via chromaticity diffusion.
\newblock {\em IEEE Transactions on Image Processing}, 10(5):701--707, 2001.

\bibitem{Tibshirani96}
R.~Tibshirani.
\newblock Regression shrinkage and selection via the lasso.
\newblock {\em Journal of the Royal Statistical Society B}, 58(1):267--288,
  1996.

\bibitem{WWainwright09}
M.~Wainwright.
\newblock Sharp thresholds for high-dimensional and noisy sparsity recovery
  using $\ell^{1}-$constrained quadratic programming (lasso).
\newblock {\em IEEE Transactions on Information Theory}, 55(5):2183--2202,
  2009.

\bibitem{WPMJR10}
D.~L. Woodard, S.~Pundlik, P.~Miller, R.~Jillela, and A.~Ross.
\newblock On the fusion of periocular and iris biometrics in non-ideal imagery.
\newblock In {\em IEEE 20th International Conference on Pattern Recognition
  (ICPR)}, pages 201--204, Istanbul, Turkey, 2010.

\bibitem{woodard2010}
D.L. Woodard, S.~Pundlik, P.~Miller, R.~Jillela, and A.~Ross.
\newblock On the fusion of periocular and iris biometrics in non-ideal imagery.
\newblock In {\em Pattern Recognition (ICPR), 2010 20th International
  Conference on}, pages 201 --204, August 2010.

\bibitem{WYGSM09}
J.~Wright, A.~Y. Yang, A.~Ganesh, S.~Sastry, and Y.~Ma.
\newblock Robust face recognition via sparse representation.
\newblock {\em IEEE Transactions on Pattern Analysis and Machine Intelligence},
  31(2):210--227, 2009.

\bibitem{WStiles82}
G.~Wyszecki and W.~Stiles.
\newblock {\em Color Science: Concepts and Methods, Quantitative Data and
  Formulas}.
\newblock Wiley, New York, NY, USA, 1982.

\bibitem{HZou06}
H.~Zou.
\newblock The adaptive lasso and its oracle properties.
\newblock {\em Journal of the American Statistical Association},
  101(476):1418--1429, 2006.

\bibitem{ZHastie05}
H.~Zou and T.~Hastie.
\newblock Regularization and variable selection via the elastic net.
\newblock {\em Journal of the Royal Statistical Society: Series B},
  67(2):301--320, 2005.

\bibitem{ZZhang09}
H.~Zou and H.~Zhang.
\newblock On the adaptive elastic-net with a diverging number of parameters.
\newblock {\em The Annals of Statistics}, 37(4):1733--1751, 2009.

\end{thebibliography}
\end{document}